\documentclass[a4paper,12pt,headsepline]{scrreprt}

\usepackage[utf8]{inputenc}

\usepackage{graphicx}

\usepackage[english]{babel}

\usepackage[right]{eurosym}

\usepackage[T1]{fontenc}

\usepackage{lmodern}

\usepackage{longtable}

\usepackage{geometry}
\usepackage{fancybox}

\usepackage[hyphens,obeyspaces,spaces]{url}

\usepackage{color}

\usepackage{amssymb}
\usepackage{bm}
\newcommand{\titleDocument}{Master Thesis}
\newcommand{\subjectDocument}{in physics}
\usepackage[bookmarksnumbered,pdftitle={\titleDocument},hyperfootnotes=false]{hyperref}

\usepackage{array}

\usepackage{setspace}

\usepackage{capt-of}

\usepackage{makeidx}

\usepackage{listings}
\makeindex

\usepackage[english]{nomencl}

\makenomenclature

\usepackage{shellesc}
\usepackage{amsmath}
\usepackage{amssymb}

\usepackage{bbold}
\usepackage{siunitx}
\usepackage{booktabs}
\usepackage{caption}
\usepackage[backend=biber,natbib=true,
url=false, 
doi=true,
eprint=true]{biblatex}
\usepackage{csquotes}
\usepackage{float}
\usepackage{subcaption}
\usepackage{amsthm}
\usepackage[final]{pdfpages}

\usepackage{listings}
\usepackage{xcolor}

\definecolor{codegreen}{rgb}{0,0.6,0}
\definecolor{codegray}{rgb}{0.5,0.5,0.5}
\definecolor{codepurple}{rgb}{0.58,0,0.82}
\definecolor{backcolour}{rgb}{0.95,0.95,0.92}

\lstdefinestyle{mystyle}{
    backgroundcolor=\color{backcolour},   
    commentstyle=\color{codegreen},
    keywordstyle=\color{magenta},
    numberstyle=\tiny\color{codegray},
    stringstyle=\color{codepurple},
    basicstyle=\ttfamily\footnotesize,
    breakatwhitespace=false,         
    breaklines=true,                 
    captionpos=b,                    
    keepspaces=true,                 
    numbers=left,                    
    numbersep=5pt,                  
    showspaces=false,                
    showstringspaces=false,
    showtabs=false,                  
    tabsize=2
}

\usepackage{multirow}
\begin{document}

\thispagestyle{empty}

\begin{verbatim}


\end{verbatim}

\begin{center}
\Large{Universität Leipzig}\\
\end{center}

\begin{center}
\Large{Fakultät für Physik und Geowissenschaften}\\
\Large{Institut für Theoretische Physik}
\end{center}

\begin{verbatim}




\end{verbatim}
\begin{center}
\doublespacing
\textbf{\LARGE{\titleDocument}}\\
\singlespacing
\begin{verbatim}

\end{verbatim}
\textbf{{~\subjectDocument~}}
\end{center}
\begin{verbatim}

\end{verbatim}
\begin{center}

\end{center}
\begin{verbatim}

\end{verbatim}
\begin{center}
\textbf{for the attainment of the academic degree \\ Master of Science}
\end{center}
\begin{verbatim}






\end{verbatim}
\begin{flushleft}
\begin{tabular}{llll}
\textbf{Topic:} & &Hessian Eigenvectors and Principal Component Analysis& \\
&  & of Neural Network Weight Matrices&\\
& & &\\
\textbf{Author:} & & David Haink& \\
& & &\\
\textbf{Version from:} & & September 25, 2023 &\\
& & &\\
\textbf{1st Supervisor:} & & Prof. Dr. Bernd Rosenow&\\
\textbf{2nd Supervisor:} & & Dr. Matthias Thamm&\\
\end{tabular}
\end{flushleft}

\pagenumbering{roman}

\onehalfspacing

\newpage

\thispagestyle{empty}

\section*{Abstract}
This study delves into the intricate dynamics of trained deep neural networks and their relationships with network parameters. Trained networks predominantly continue training in a single direction, known as the drift mode. This drift mode can be explained by the quadratic potential model of the loss function, suggesting a slow exponential decay towards the potential minima.
We unveil a correlation between Hessian eigenvectors and network weights. This relationship, hinging on the magnitude of eigenvalues, allows us to discern parameter directions within the network. Notably, the significance of these directions relies on two defining attributes: the curvature of their potential wells (indicated by the magnitude of Hessian eigenvalues) and their alignment with the weight vectors.
Our exploration extends to the decomposition of weight matrices through singular value decomposition. This approach proves practical in identifying critical directions within the Hessian, considering both their magnitude and curvature. Furthermore, our examination showcases the applicability of principal component analysis in approximating the Hessian, with update parameters emerging as a superior choice over weights for this purpose.
Remarkably, our findings unveil a similarity between the largest Hessian eigenvalues of individual layers and the entire network. Notably, higher eigenvalues are concentrated more in deeper layers.
Leveraging these insights, we venture into addressing catastrophic forgetting, a challenge of neural networks when learning new tasks while retaining knowledge from previous ones. By applying our discoveries, we formulate an effective strategy to mitigate catastrophic forgetting, offering a possible solution that can be applied to networks of varying scales, including larger architectures.
This study is a step to uncover intricate behavior of deep neural networks and also provides practical solutions for enhancing their capabilities and addressing critical challenges.

\newpage

\thispagestyle{empty}
\section*{Acknowledgements}
I would like to express my heartfelt gratitude to several individuals who have played instrumental roles in the completion of this thesis.
First and foremost, I extend my sincere appreciation to Marcel Kühn, Matthias Thamm, and Max Staats. Their invaluable insights, thoughtful discussions, and unwavering support have been pivotal throughout this research journey. Without their guidance, this thesis would not have been possible.
I am also deeply thankful to Professor Dr. Bernd Rosenow for his mentorship, and continuous encouragement. His expertise has been a guiding light in navigating the complexities of this study.
Furthermore, I extend my appreciation to Mr. Thamm and Mr. Staats for their dedicated supervision, which has contributed significantly to the refinement of this work.

\singlespacing

\newpage
\phantomsection

\setcounter{page}{1}

\thispagestyle{empty}
\addcontentsline{toc}{section}{Table of contents}
\tableofcontents

\newpage
\thispagestyle{empty}
\mbox{}
\newpage

\onehalfspacing

\pagenumbering{arabic}

\chapter{Introduction}
\label{Introduction}
Gottfried Wilhelm Leibniz foresaw the potential of solving complex mathematical problems through binary classifications of True or False, thus laying the groundwork for the first mechanical computer \cite{Schmidhuber}. As time passed, it became increasingly clear that computers easily surpassed humans in tasks that can be broken down into straightforward algorithms, consisting of a simple list of instructions. The key issue is that any problem we can formulate into an algorithm can be solved, given enough computational power and processing time. However, a significant challenge arose with this advancement: How to overcome obstacles that do not easily translate into algorithmic structures? Take, for example, the field of image classification. The task of differentiating between inanimate objects and living animals is a relatively straightforward one. This can be achieved by developing guidelines to check for the existence of specific attributes like eyes, noses, and fur, which are unique to animals and not present in stones. However, the task becomes more intricate when faced with the challenge of differentiating between cats and dogs. While dogs and cats share many similarities, it is difficult to pinpoint the exact distinguishing features that universally set all dogs apart from all cats.
In response to this enigma, a bold solution emerged from an unexpected source: The process of translating these complex challenges into algorithmic frameworks, akin to learning, underwent a transformational shift, becoming an algorithm on its own that can be resolved computationally \cite{ivakhnenko1967cybernetics}. By digitizing data and the desired solution for these enigmas, we applied the fundamental principles of derivatives - attributable to Leibniz's legacy \cite{Schmidhuber} - to autonomously simplify these complexities. For example, when analyzing pictures, we supply the computer with images of both dogs and cats and allow it to determine a way to differentiate between the two. This process encompasses various names, initially as Cybernetics, later evolving into Artificial Intelligence, and currently maturing as Deep Learning.\\
The principles underpinning Deep Learning were not exclusively forged within the realm of scientific inquiry, but rather stem from the very fabric of nature itself. Our cognitive architecture, constituted by neurons intricately interconnected, forms a neural network that embodies the essence of learning. Analyzing the structure and functioning of our brain enabled us to decipher the algorithmic mechanisms of learning, which could then be harnessed and adapted to meet the computational requirements of our machines \cite{rosenblatt1958perceptron}. As a consequence, the learning algorithms we employ bear the moniker of Deep Neural Networks (DNNs), a nomenclature that resonates with the intricate neural structure and functionality of our own minds. Yet, the profound irony remains: Just as we sometimes find it challenging to elucidate how our own brain navigates certain problems, providing the same explanations to our computational counterparts, as in the case of distinguishing between cats and dogs, proves to be equally formidable.\\
In recent years, DNNs have emerged as a transformative force across diverse domains, achieving remarkable feats in various applications such as natural language translation like DeepL and Google Translator, conversational agents like ChatGPT and Bing Chat, and strategic game-playing in chess and go \cite{silver2016mastering}. These networks' extraordinary capabilities underscore their potential to reshape our technological landscape fundamentally. Yet, beneath their impressive performance lies a veil of mystery, as the decisions made by DNNs often remain opaque and elusive. Since we no longer can comprehend the instructions the computer finds, when searching for an algorithm \cite{HEUILLET2021106685}.\\
The intrinsic power of DNNs originates from their ability to model intricate relationships within complex data, converting raw inputs into meaningful predictions. These models, characterized by webs of interconnected parameters, raise fundamental questions about the interplay between network architecture, training dynamics, and task proficiency. Addressing these questions not only holds theoretical significance but also carries practical implications for enhancing the reliability, interpretability, and generalization of DNNs in real-world applications \cite{bengio2017deep}.\\
Driven by the quest to unveil the inner workings of DNNs, researchers have pursued various approaches, each aimed at shedding light on distinct aspects of these models. Two of many paradigms warrant our attention: Random Matrix Theory (RMT) and Principal Component Analysis (PCA) of training dynamics. These complementary strategies offer unique insights into the behavior of DNNs and the factors driving their performance.\\
RMT, rooted in the mathematical foundations of linear algebra, delves into the structure of DNNs' weight matrices. These matrices, emerging from the network's layered architecture and initialized with randomness, undergo change during training, giving rise to an interplay between deterministic and random components. RMT together with singular value decomposition can be used to discern meaningful patterns within these matrices, shedding light on the evolution of network parameters and identifying key components that drive the network's performance \cite{DBLP:journals/corr/abs-1810-01075}.\\
In contrast, PCA of training dynamics delves into the trajectory of network updates as dictated by the Stochastic Gradient Descent (SGD), the workhorse optimization algorithm driving DNN training. By analyzing the development of parameter updates, this approach offers a window into the network's adaptation process and unveils directions of maximal change, which are intrinsically tied to critical information for task-solving \cite{doi:10.1073/pnas.2015617118}.\\
While both RMT and PCA have independently contributed to advancing our understanding of DNNs, a comprehensive comparative exploration that compares these approaches remains absent. Bridging the gap between the network parameters and training dynamics holds the potential to catalyze a more comprehensive behavior of DNN and yield insights with far-reaching ramifications. Beyond theoretical implications, these insights could pave the way for practical applications, such as combatting catastrophic forgetting \cite{doi:10.1073/pnas.2015617118} and navigating the challenges posed by noisy data \cite{Staats.2022}.\\
In light of these considerations, this study embarks on an journey to untangle DNN behavior by synergistically employing RMT and PCA. Through a comparative analysis, we strive to elucidate the interplay between network architecture, training dynamics, and task performance. Our exploration not only enriches the theoretical discourse surrounding DNNs but also holds the promise of empowering practical solutions for the challenges that lie ahead.\\
The subsequent chapters of this thesis are structured as follows: Chapter 2 introduces the theoretical and mathematical foundations of DNNs necessary for comprehending the subsequent sections. Chapter 3 explores PCA and its current state of research in the context of DNNs. Chapter 4 delves into RMT and its application to DNNs. Chapter 5 provides insights into the practical aspects of network realization, including network setups and datasets used. Chapter 6 presents our comparative analysis of PCA and RMT results in DNNs, along with an exploration of PCA properties in relation to the Hessian matrix. Chapter 7 shifts our focus to the Hessian matrix, its properties, its relationships with network weights, and its connection to RMT. Finally, we discuss our findings, offer suggestions for further research, and conclude this work.

\chapter*{Notation}
Throughout this thesis, vectors are represented in lowercase bold, and higher-order tensors are represented in uppercase bold. Fundamental mathematical theorems and physical laws are assumed to be known and will only be referenced when applied in the context of this work.

\chapter{Introduction to Deep Neural Networks}
This chapter delves into modern Deep Neural Networks (DNNs) for image recognition \cite{lecun1995convolutional,krizhevsky2017imagenet,simonyan2014very, he2016deep}, which are designed to transform inputs, denoted as $\bm{x}$ (e.g., images), into outputs, represented by $\bm{y}$ (e.g., image labels), in order to classify images. Formally, the relationship is expressed as $\bm{y} = f(\bm{x},\bm{w})$, where $\bm{w}$ signifies the network's weights. Deep Neural Networks are structured in layers, which can be interpreted as transformations themselves.
Let $\bm{y}^{(l)}=f^{(l)}(\bm{x}^{(l)},\bm{w}^{(l)})$ be the output of layer $l$. The weights of a layer are a part of all weights. The input of a layer is commonly the output of the previous layer, i.e. $\bm{x}^{(l)}=\bm{y}^{(l-1)}$ \cite{bengio2017deep}.

\section*{Activation Functions}
Activation functions modify the output of each layer to introduce nonlinearity. A widely used activation function is the Rectified Linear Unit (ReLU) \cite{glorot2011deep}:
\begin{equation}
	y_i(\bm{x}) = \max(x_i, 0)\, .
\end{equation}
Another important activation function is the softmax function:
\begin{equation}
	\label{softmax}
	y_i(\bm{x}) = \frac{e^{x_i}}{\sum_je^{x_j}}\, .
\end{equation}
The softmax activation is commonly employed in the output layer for classification tasks, providing probabilities for each class. The class with the highest probability is the predicted label, often referred to as the top-1 accuracy.
\section*{Loss Function}
\label{sec:loss}
The network's performance assessment and enhancement are facilitated through a loss function, typically employing the categorical cross-entropy loss:
\begin{equation}
	L(\bm{x},\bm{z}) = -\sum_iz_i\log(y_i(\bm{x}))\, ,
\end{equation}
where $z_i = 1$ only for the correct output label of input $\bm{x}$ and $0$ otherwise. This loss function is conceptually aligned with information entropy, rewarding accurate approximations of the training data's probability distribution.
\section*{Stochastic Gradient Descent}
\label{sec:sgd}
Let us consider datasets of inputs $\bm{x}^i$ and their corresponding correct labels $\bm{z}^i$, where the upper index corresponds to a single sample of the dataset.
When training the network, the loss function is computed for multiple training samples in a batch $\mathcal{B}_k\equiv\mathcal{B}_{k(t)}$:
\begin{equation}
	l^{(k)}(\bm{x},\bm{z})=\frac{1}{S}\sum_{i\in\mathcal{B}_k} L(\bm{x}^{(i)},\bm{z}^{(i)})\, .
\end{equation}
Batches are typically of size $S\in[32,512]$ \cite{2107.09133}.
The network updates its weights using Stochastic Gradient Descent (SGD) $\bm{w}(t+1)=\bm{w}(t)+\bm{v}(t+1)$, where $t$ is a time step and:
\begin{equation}
	\label{eq:langevin}
	\bm{v}(t+1) = -\eta\nabla_{\bm{w}}l^{(k(t))}(\bm{w},\bm{x},\bm{z}) + \beta\bm{v}(t)\, ,
\end{equation}
where $\eta > 0$ is the learning rate and $\beta \in [0,1)$ is the momentum. The term $\nabla_{\bm{w}}l^{(k(t))}(\bm{w},\bm{x},\bm{z})$ corresponds to the gradient of the loss for a batch $\mathcal{B}_k(t)$ with respect to the network's weights. The concept of stochasticity arises from the random sampling of batches and the variability introduced by this process.\\
A challenge that arises due to updating all weights from the last layer backwards is the vanishing gradient problem, wherein the gradients diminish in magnitude as the optimization process progresses, particularly affecting early layers, due to the chain rule of derivatives. This phenomenon hinders the learning of these layers and can be attributed to the multiplication of gradients during the chain rule process \cite{hochreiter2001gradient}.
\section*{Regularization}
Regularization techniques play a critical role in preventing overfitting. One prevalent form is the $L_2$ regularization, adding $\lambda||\bm{w}||^2_2$ to the loss $l^{(k)}$ in Eq.\eqref{eq:langevin}. This technique discourages the network from focusing excessively on unimportant features, improving the model's generalization capability.
\section*{Layers}
Various layer types are employed in contemporary DNN architectures \cite{bengio2017deep}, including convolutional layers, pooling layers, and dense layers. These layer types are mathematically characterized below.
\subsection*{Dense Layers}
Dense layers implement linear matrix multiplications on flattened inputs:
\begin{equation}
	\bm{y} = \bm{W}\bm{x} + \bm{b}\, ,
\end{equation}
where $\bm{y}$ is the output, $\bm{W}$ is the weight matrix, $\bm{x}$ is the input, and $\bm{b}$ is the bias vector. Flattening is applied to matrix inputs, converting them into vectors. These layers form a versatile foundation by linking each input to each output individually \cite{lecun2015deep}.
\subsection*{Convolutional Layers}
3-D convolutional layers involve convolutions of input $\bm{X}$ with a weight tensor $\bm{W}$:
\begin{equation}
	Y_{ijk} = \sum_{m\in I_1}\sum_{n \in I_2}\sum_{l \in I_3} W_{mnlk}X_{i+m,j+n,l}+B_{ijk}\, ,
\end{equation}
where $I_1=[-d_1,d_1]$, $I_2=[-d_2,d_2]$, $I_3=[1,d_3]$. $d_1$ and $d_2$ are the kernel sizes in the x and y directions, respectively, and $d_3$ corresponds to the input depth. These layers are particularly efficient for image recognition tasks, as they exploit local correlations among nearby pixels \cite{lecun2015deep}.
\subsection*{Pooling Layers}
Max Pooling layers, often employed after convolutional layers, increase translation invariance. They select the maximum value from neighboring inputs within a translational distance $d$:
\begin{equation}
	Y_{ij} = \max_{m,n \in K}{X_{i+m,j+n}}, \ K=[-d,d]\, .
\end{equation}
Max Pooling enhances robustness by retaining key features while reducing sensitivity to small input variations \cite{bengio2017deep}.
\subsection*{Batch Normalization Layer}
Batch normalization normalizes the input according to the rule:
\begin{equation}
	\label{eq:batchnorm}
	\bm{y} = \gamma \frac{\bm{x}-\langle\bm{x}\rangle_k}{\sigma^2_k(\bm{x})+\epsilon}+\beta\, ,
\end{equation}
where $\gamma=1$, $\beta=0$ are learnable parameters, $\epsilon$ is a small constant, $\langle\bm{x}\rangle_k$ is the mean of the input batch, and $\sigma^2_k$ is its variance. The layer enhances training speed and testing accuracy by aligning input statistics with a standard distribution \cite{ioffe2015batch}.

\subsection*{Residual Layers}
In very deep networks, it has been found useful to include shortcuts, called residuals, in the network to further increase the accuracy of the test data.
If we denote one of the previously defined layers or multiple layers as a transformation, i.e., $\bm{y}^{(l)}=f^{(l)}(\bm{x}^{(l)})$, then the shortcut is as simple as:
\begin{equation}
	\bm{y}^{(l)}=f^{(l)}(\bm{x}^{(l)})+\bm{x}^{(l)}\, .
\end{equation}
For the residual to be well-defined, it is required that the dimension of $\bm{x}^{(l)}$ and $\bm{y}^{(l)}$ match, or that we can propagate the input to match the dimension of the output, e.g. when adding to a higher order convolutional layer. While residual networks proved to generalize better than those without shortcuts and without increasing the number of parameters, computing the gradient of batches becomes more costly by increasing the number of floating point operations (FLOP) the network needs to perform greatly \cite{he2016deep}.

\chapter{Dynamics in Deep Neural Networks}
\label{sec:PCA}
\section{The Loss Landscape and the Hessian Matrix}
To comprehend network dynamics, it is useful to consider the Hessian matrix:
\begin{equation}
	H_{ij}=\frac{\partial^2 L(\bm{w})}{\partial{w_i}\partial{w_j}}
\end{equation}
Several studies \cite{doi:10.1073/pnas.2015617118,2107.09133} propose a theoretical similarity between eigenvectors of the Hessian matrix and the covariance matrix of mean gradients. This similarity is observed in successful networks with a high test accuracy.
Following \cite{jastrzębski2018factors}, it is posited that: 
\begin{equation}
	\label{eq:loss_hess_quad}
	L(\bm{w})\approx L_{\text{min}}+(\bm{w}-\bm{\mu})^T\bm{\frac{H}{2}}(\bm{w}-\bm{\mu})\, ,
\end{equation}
where $L_{\text{min}}$ represents the minimum of the loss function, and $\bm{\mu}$ denotes the coordinates of the loss function's minimum in weight space.
\section{Principal Component Analysis}
Consider a weight matrix flattened to a vector, denoted as $\bm{w}(t)$, for a specific layer or the entire network, measured over a sequence of $T\geq n$ discrete time steps $t$ such that $[0,T]\subset\mathbb{N}_0\rightarrow \mathbb{R}^n$. The covariance matrix can be defined as:
\begin{equation}
	\Sigma_{ij} = \langle w_i w_j-\langle w_i\rangle \langle w_j\rangle\rangle\, ,
\end{equation}
where $\langle X(t) \rangle=\frac{1}{T}\sum^{T}_{t=0} X(t)$. The covariance matrix quantifies the relationships between different components of the weight vector over time.
The eigenvectors of the covariance matrix are termed principal components, represented as $\bm{p}_i$. Corresponding to these eigenvectors are their associated eigenvalues, denoted as $\sigma^2_i$:
\begin{equation}
	\label{eigvalpca}
	\bm{\Sigma}\bm{p}_i = \sigma^2_i\bm{p}_i\, .
\end{equation}
The eigenvectors furnish an orthonormal basis for the $\mathbb{R}^n$ space, given the symmetry of $\bm{\Sigma}$. Consequently, it becomes possible to express the weight matrix in this basis:
\begin{equation}
	\bm{w}(t) = \sum^{n}_{i=1}\theta_i(t)\bm{p}_i, \quad \theta_i(t) := \bm{w}(t)\cdot\bm{p}_i\, .
\end{equation}
It is noteworthy that the variance of $\theta(t)$ corresponds to the eigenvalue of the respective principal component.
Using a decomposition into principal components allows to examine the dynamics of weights in a late training phase known as the exploration phase \cite{doi:10.1073/pnas.2015617118}. During this phase, the generalization error does not improve significantly, and the training loss function changes gradually. The central notion is that the network is proximate to a minimum of $L(\bm{w})$, and further training does not cause it to deviate substantially from this minimum.
When the weights vary in the direction of the principal components, the loss function can be represented as:
\begin{equation}
	L(\delta\theta)_i:=L(\bm{w}+\delta\theta\bm{p}_i)\, .
\end{equation}
This behavior is observed in \cite{doi:10.1073/pnas.2015617118} to follow a potential well:
\begin{equation}
L(\delta\theta)_i \propto \delta\theta^2
\end{equation}
as $\delta\theta\rightarrow 0$. The eigendirections of the loss landscape may not align precisely with the principal components. Nonetheless, this loss function behavior holds true in all directions spanned by eigenvectors of the Hessian matrix with sufficiently large eigenvalues.
Due to the observation of a quadratic potential well and by that the principal components and Hessian eigenbasis are supposedly close in being diagonal as shown in \cite{jastrzębski2018factors}, the flatness $F_i$ of the minima of the principal components and the Hessian eigenvalues can be related to $F_i^{-2}\propto h_i$.
The eigenvalues of the principal components can then be linked to the eigenvalues of the Hessian matrix:
\begin{equation}
	\label{eq:inverse_flatness}
	\sigma^2_i\propto h^\alpha_i\, ,
\end{equation}
where $\sigma^2_i$ are the eigenvalues of the principal components in descending order, and $h_i$ are the Hessian eigenvalues in descending order. Experimental findings of \cite{doi:10.1073/pnas.2015617118} suggest that $\alpha \approx 2$, shown only for the flatness. In particular, \cite{kühn2023correlated} shows that the empirical result of $\alpha$ is affected by measuring for only a short period of time and provides a theoretical relationship:
\begin{equation}
    \sigma^2_i\propto
    \begin{cases}
		h_i& h_i < h_\text{cross}\\
        \text{const}&h_i>h_\text{cross}\, ,
	\end{cases}
\end{equation}
where $h_\text{cross}:=3S\frac{1-\beta}{\eta N_\text{train}}$, S is the batch size, $\beta$ signifies the momentum, and $N_\text{train}$ is the number of training examples per epoch. The dependency of $\alpha$ on the corresponding eigenvalue highlights its non-constant nature.

\section{Catastrophic Forgetting}
\label{sec:cata_forget}
Catastrophic forgetting emerges when a network learns multiple tasks independently, resulting in performance degradation on previously learned tasks. This issue stems from optimizing the loss function for one task, inadvertently neglecting others, and allowing their associated loss values to escalate.
To mitigate this phenomenon, \cite{doi:10.1073/pnas.2015617118} suggests weakly constraining the largest $N_\text{lim}$ principal components through a regularization term integrated into the loss function:
\begin{equation}
    L_\text{cf}(\bm{w})=\lambda_{cf}\sum^{N_\text{lim}}_{i=1}\frac{1}{F^2_i}\left((\bm{w}-\tilde{\bm{\mu}}_1)\cdot\bm{p}_i\right)^2\, ,
\end{equation}
where $\lambda_{cf}$ denotes a positive real regularization constant, $\tilde{\bm{\mu}}_1$ represents the weights of the network or layer at the end of training for the previous task that are close to a minimum $\bm{\mu}_1$, and $F_i$ signifies the flatness in direction of the principal components, while $\bm{p}_i$ corresponds to their associated principal component, computed after training for the previous task with the dataset of the previous task.
It is noteworthy that due to Eq.\eqref{eq:inverse_flatness}, measuring $N_\text{lim}+1$ time steps is deemed sufficient to approximate the $N_\text{lim}$ the largest Hessian eigenvectors using principal components, rendering computation more feasible, if we assume that the principal components approximate the Hessian eigenvectors sufficiently. In \cite{doi:10.1073/pnas.2015617118} this relation was applied to connect the regularization to the Hessian landscape model. We can use this relation backwards to convert the problem to one of the Hessian:
\begin{equation}
    L_\text{cf}(\bm{w})=\lambda_{cf}\sum^{N_\text{lim}}_{i=1}h_i\left((\bm{w}-\tilde{\bm{\mu}}_1)\cdot\bm{h}_i\right)^2\, ,
\end{equation}
where again $\lambda_{cf}$ denotes a positive real regularization constant, $\tilde{\bm{\mu}}_1$ represents the weights of the network or layer at the end of training for the previous task that are close to a minimum $\bm{\mu}_1$, and $h_i$ signifies the Hessian eigenvalues in descending order, while $\bm{h}_i$ corresponds to their associated eigenvector, computed after training for the previous task with the dataset of the previous task. When all eigenvalues are considered, this equation exhibits similarity to Eq.\eqref{eq:loss_hess_quad}.

\chapter{Random Matrix Theory and Singular Value Decomposition in Deep Neural Networks}
\label{sec:RMT}
The weight matrix $\bm{W}\in \mathbb{R}^{N\times M}$ of a given layer can be decomposed using singular value decomposition (SVD):
\begin{equation}
	\bm{W}=\bm{U}\bm{\Sigma}\bm{V}^T, \quad \bm{\Sigma}=\text{diag}(\nu_1,...,\nu_N)\, ,
\end{equation}
where $\bm{U}$ and $\bm{V}$ are orthogonal matrices and $\nu_1,...,\nu_N$ are the singular values, arranged in decreasing order.
These singular values are accompanied by their corresponding right singular vectors, which form the rows of $\bm{V}$, and the left singular vectors, which are extracted from $\bm{U}$.
When elements of the weight matrix $W_{ij}$ adhere to a normal distribution $W_{ij}\sim \mathcal{N}(0,\sigma^2_\text{mp})$, an observation can be made for the limit ${N,M\rightarrow \infty}$ with $Q:=N/M\in\mathbb{R}_{\geq 1}$, which gives rise to a Marchenko-Pastur distribution (MP):
\begin{equation}
	\label{eq:MP}
	\rho(\nu) =
	\begin{cases}
		\frac{Q}{2\pi\sigma^2_\text{mp}\lambda}\sqrt{(\lambda^+-\lambda)(\lambda-\lambda^-)}\  & \text{if}\ \lambda\in[\lambda^-,\lambda^+] \\
		0                                                                                      & \text{otherwise}\, ,
	\end{cases}
\end{equation}
where $\lambda^\pm=\sigma^2_\text{mp}\left(1\pm\frac{1}{\sqrt{Q}}\right)^2$ \cite{DBLP:journals/corr/abs-1810-01075}.
For deep neural networks, it has been demonstrated that the behavior of singular values mirrors that of random matrices \cite{DBLP:journals/corr/abs-1810-01075}. Specifically, they adhere to the Marchenko-Pastur distribution (Eq.\eqref{eq:MP}). This behavior is attributed to their initialization as randomly distributed values, and due to the fact that the majority of weights undergo limited change during training.
Among the scrutinized trained networks, a few singular values reside outside the bulk and encapsulate nearly all network information. This is highlighted by the fact that removing bulk singular values by setting them to zero does not detrimentally affect network accuracy \cite{Staats.2022}.
Singular values hold a close relationship with the eigenvalues of the PCA applied to the same matrix. The eigenvalues obtained from PCA are essentially the square of the singular values. In light of this connection, the distributions of the unfolded spacings between singular values exhibit a behavior akin to the Wigner surmise, which the RMT theory predicts. This surmise is captured by the equation:
\begin{equation}
	p(s)=\frac{\pi s}{2}e^{-\frac{\pi s^2}{4}}\, ,
\end{equation}
where $s=\xi_{n}-\xi_{n+1}$ is the unfolded spacing, with $\xi_n$ are the singular values, such that the spacings are locally normalized \cite{livan2018introduction}.
In a recent study \cite{Thamm_2022}, compelling evidence was presented demonstrating that the singular values of weight matrices in DNNs conform to the expectations set forth by the Wigner surmise.

\chapter{Deep Neural Network Setup}
\section{Datasets}
The primary dataset for comparison is CIFAR-10 \cite{krizhevsky2009learning}, which consists of 10 classes including airplanes, cars, birds, cats, deer, dogs, frogs, horses, ships, and trucks. Each image has $32\times32$ pixels and three color values for red, green, and blue (RGB) from 0 to 255 (uint8), so each image is a tensor of shape $(32,32,3)$. The dataset comprises 60000 images, with 6000 images per class. A validation set of 10000 images is used to estimate the generalization error.
While CIFAR-10 serves as a benchmark for smaller networks, more modern, larger networks tend to achieve test accuracies beyond $99.5\%$ \cite{kabir2023reduction}, making comparisons challenging due to the high accuracy ceiling.\\
For more complex networks, ImageNet \cite{russakovsky2015imagenet} is a significant benchmark. However, due to the large network size and long training times, it is not employed in this study. \\
Instead, a simpler dataset, MNIST \cite{lecun1998gradient}, is additionally used. It consists of 70000 gray-scale images of handwritten digits with a $28\times28$ pixel resolution. This dataset is considered to be much simpler than CIFAR-10, since even small MLP networks can perform extremely well on this dataset. Nevertheless, it is historically an important dataset, is computationally extremely easy due to the small image size, and can make analysis more feasible.
\section{Networks}
\label{sec:Networks}
\subsection*{Initialization}
\label{sec:init}
The networks' weights are initialized using random seeds. In contemporary networks, each layer is independently initialized using either a uniform or normal distribution. The Glorot initialization \cite{pmlr-v9-glorot10a}, also known as Xavier initialization, is used for uniform and normal distribution initialization.
For the uniform distribution:
\begin{equation}
\bm{W}^{(l)}\sim \mathcal{U}\left( - \sqrt{\frac{6}{n_l+n_{l+1}}},\sqrt{\frac{6}{n_l+n_{l+1}}}\right)\, ,
\end{equation} 
where $\bm{W}^{(l)}\in\mathbb{R}^{n_l\times n_{l+1}}$ represents the weights of the layer.
For the normal distribution:
\begin{equation}
\bm{W}^{(l)}\sim \mathcal{N}\left(0, \frac{2}{n_l+n_{l+1}}\right)\, .
\end{equation} 
All biases are initialized to zero. An alternative initialization used in ResNet \cite{he2016deep} does not show significant improvements over the Glorot initialization and is therefore omitted in this study.

\subsection*{Network Types}
Table \ref{tab:networks} summarizes the network types used in this study, along with their key attributes.

\begin{table}[t]
    \centering
    \begin{tabular}{c|c|c|c|c}
        Name & Dataset & \# Layers with Weights & \# Parameters & Test Accuracy\\
         \hline
         MLP 50 & MNIST & $3$ & $42k$ & $98.2\%$\\
         \hline
         MLP 256 & CIFAR-10 & $4$ & $828k$ & $44.2\%$\\
          \hline
         LeNet & CIFAR-10 & $7$ & $137k$ & $65.3\%$\\
          \hline
         miniAlexNet & CIFAR-10 & $8$ & $1023k$ & $74.9\%$\\
          \hline
         ResNet 20 & CIFAR-10 & $22$ & $272k$ & $68.1\%$
    \end{tabular}
    \caption{Summary of network types used in the study.}
    \label{tab:networks}
\end{table}

\subsubsection{MLP 50}
\label{sec:MLP50}
The Multilayer Perceptron (MLP) network consists solely of dense layers. ReLU activation is used for all layers except the last one, which employs the softmax activation function. The layer widths are $[50,50,10]$.  Fig.\ref{fig:mlp} provides a graphical representation of the network.
The layer weights are initialized with Glorot uniform.
The network is trained on the MNIST dataset for 200 epochs with a learning rate of $0.01$. The loss used is cross entropy and the update rule is SGD with a batch size of 32. We do not use momentum.
\subsubsection{MLP 256}
\begin{figure}[t]
\centering
	\begin{minipage}[t]{0.4\textwidth}
	\includegraphics[width=\textwidth,angle=0]{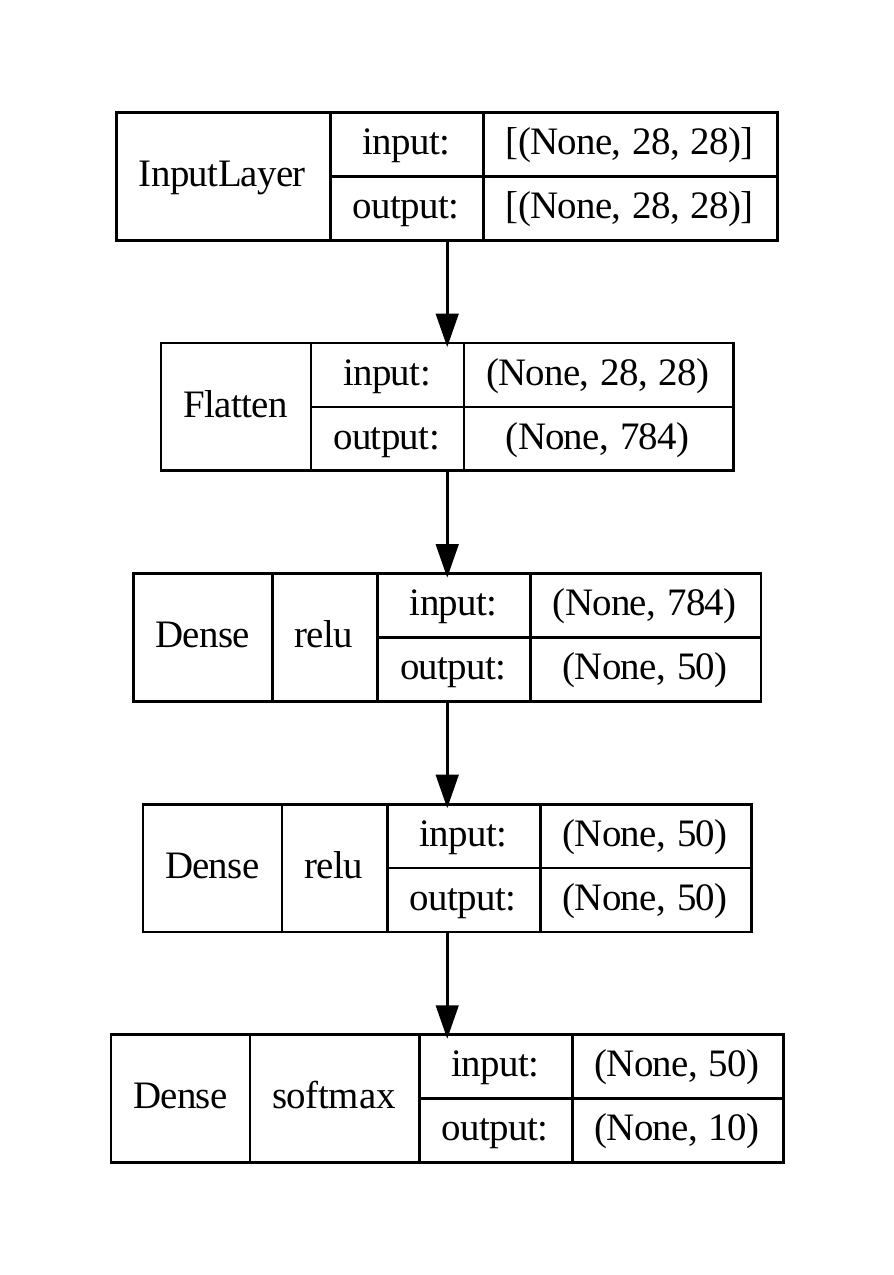}
    
     \end{minipage}
     \begin{minipage}[t]{0.4\textwidth}
    \includegraphics[width=\textwidth,angle=0]{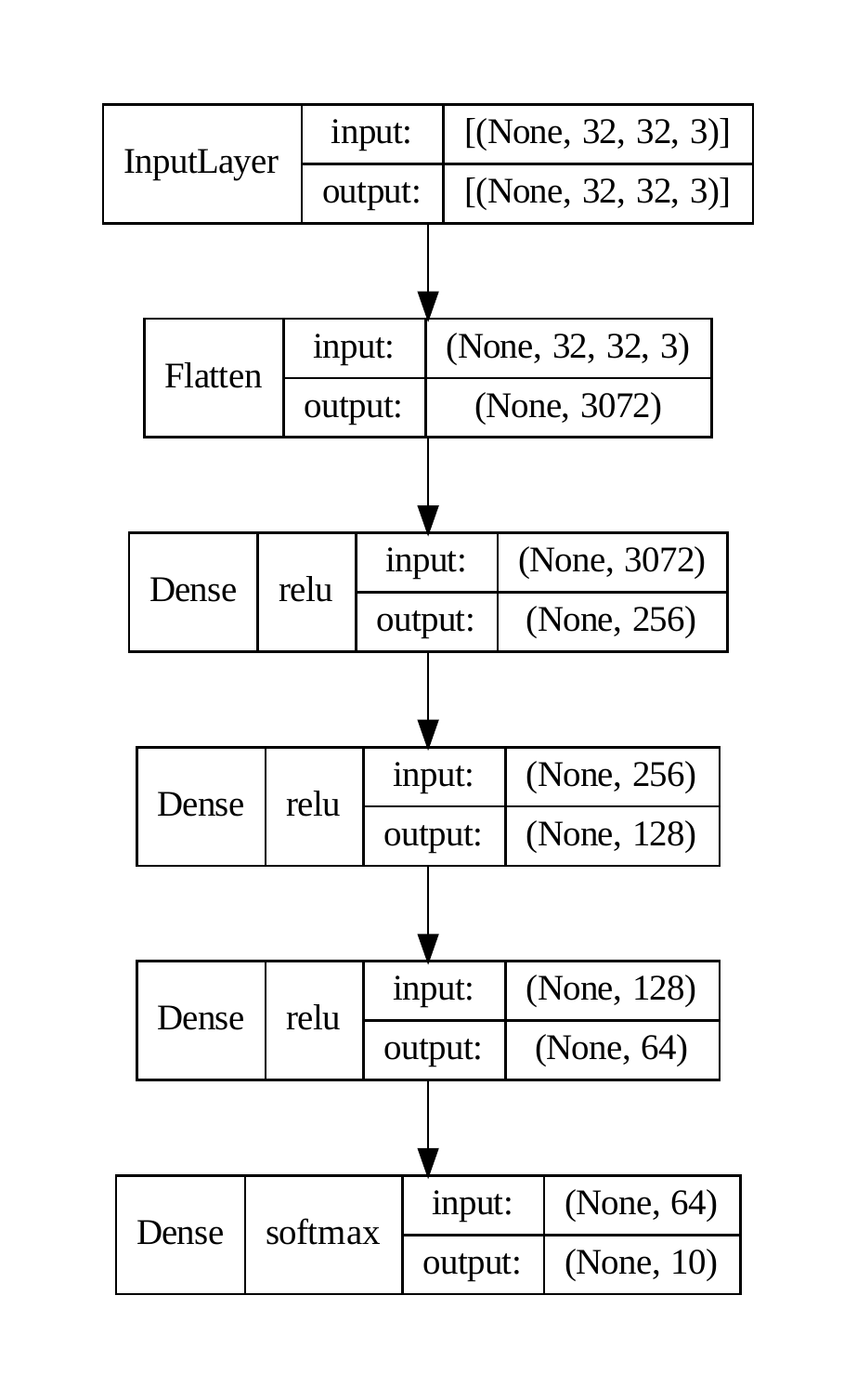}
 \end{minipage}
 
 \caption{Graphical representation of the MLP 50 (left) and the MLP 256 (right) network structure. The "None" entry means that we can insert an arbitrary batch size into the network. The input layer equal the pictures of the corresponding datasets. Those have to be turned into a vector by an additional flattening layer.}
 \label{fig:mlp}
\end{figure}

The network structure is $[256,128,64,10]$. Fig.\ref{fig:mlp} illustrates the network structure.
The layer weights are initialized with Glorot uniform. The biases are set to zero.
The network is trained on the CIFAR-10 dataset for 200 epochs with a learning rate schedule $\eta(t)=0.01\times0.99^{t},$ where the time steps $t$ are in epochs, the final learning rate is $\eta(T)\approx0.00134$. The loss used is the cross entropy and the update rule is the SGD with a batch size of 32. We do not use momentum.
To compare the differences when using weight decay or not, we regularize all layers with a weight decay constant of $5\times10^{-4}$ or, if unregularized, set the constant to zero. After training, the network achieves a full training accuracy of $100\%$ and a test accuracy of $\approx 44\%,$ depending on the seed and whether regularization is used.

\begin{figure}[t]
\centering
	\begin{minipage}[t]{0.4\textwidth}
	\includegraphics[width=\textwidth,angle=0]{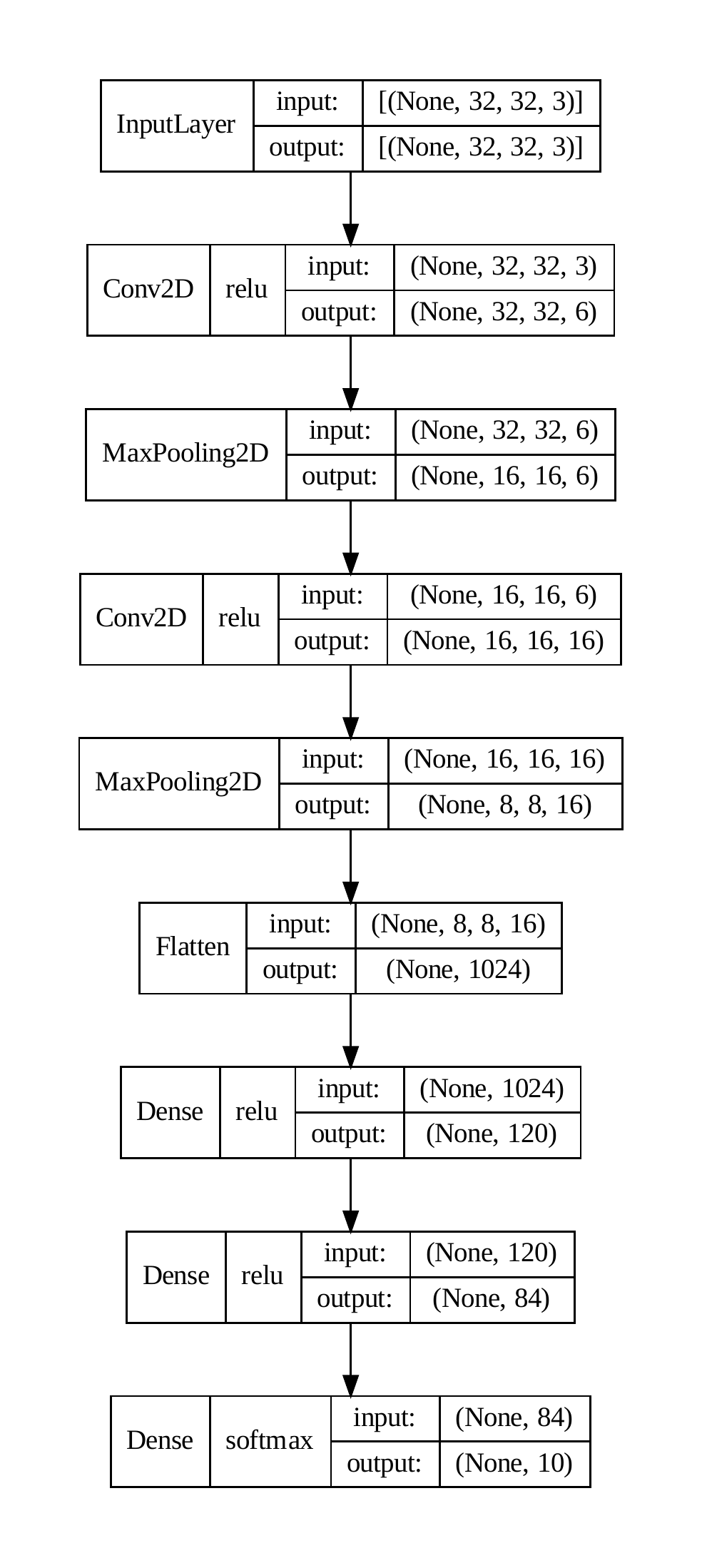}
     \end{minipage}
     \begin{minipage}[t]{0.4\textwidth}
    \includegraphics[width=\textwidth,angle=0]{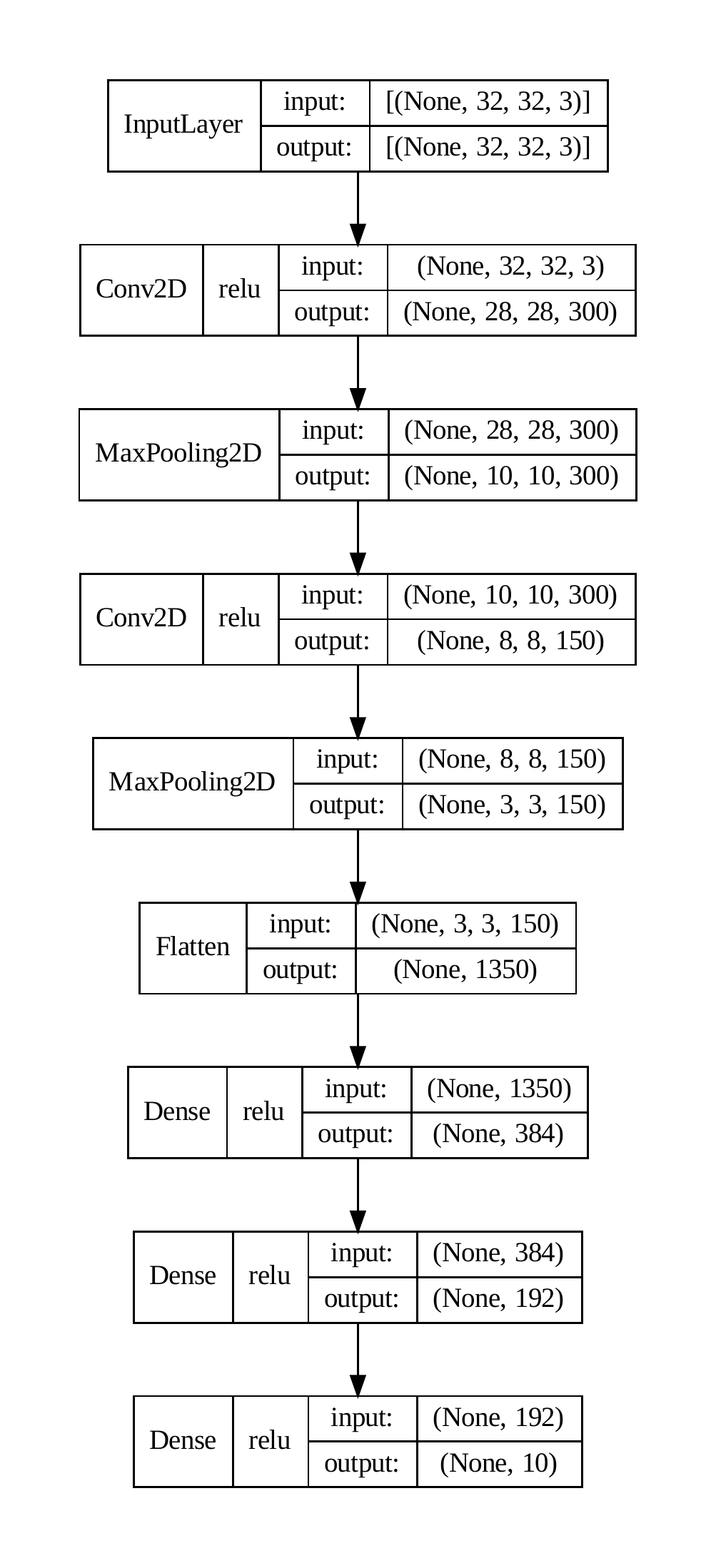}
 \end{minipage}
 \caption{Graphical representation of the LeNet (left) and miniAlexNet (right) structure. The "None" entry means that we can insert an arbitrary batch size into the network.}
 \label{fig:LeNet}
\end{figure}

\subsubsection{LeNet}
The LeNet network from \cite{lecun1998gradient} is a convolutional network optimized for the MNIST dataset. It can also be used for the CIFAR-10 dataset. The first layer is a convolutional layer with six filters and a kernel size of $(5,5)$, where the kernel is padded to preserve the shape of the input. That is, we sum the 5 closest input entries together and that we add zeros at the edges, such that we can still pad there. The next layer is a pooling layer with kernel size $(2,2)$, where the kernel is padded so that the full kernel lies in the input matrix, resulting in a smaller output dimension. This is followed by a second convolutional layer with 16 filters and a kernel size of $(5,5)$. This is followed by a pooling layer with the same properties as before. Next, the outputs are flattened into dense layers of shape $[120,84,10]$. In Fig.\ref{fig:LeNet} we can see a graphical representation of the network structure.
All layers except the pooling layers, the flattening layer, and the last layer use a ReLU activation function. The final layer has a softmax activation function.
Layer weights are initialized with Glorot normal, and biases are set to zero.
The network is trained on the CIFAR-10 dataset for 100 epochs with a learning rate schedule $\eta(t)=0.005\times0.98^{t},$ where the time steps $t$ are in epochs, the final learning rate is $\eta(T)\approx0.00134$. The loss used is the cross entropy and the update rule is the SGD with a batch size of 64. We use a momentum of $\beta=0.9$.
All layers are regularized with a weight decay of $\lambda=0.0001$.
After training, the network reaches a full training accuracy of $100\%$ and a test accuracy of $\approx 65.3\%,$ depending on the seed.
\subsubsection{miniAlexNet}
AlexNet \cite{krizhevsky2017imagenet} is a convolutional network similar to LeNet, but has many more parameters and is optimized for the ImageNet dataset. Its smaller variant, the miniAlexNet or in \cite{zhang2017understanding} called small AlexNet, is instead optimized for CIFAR-10 and is used in this work because the original AlexNet is too large for our analysis.
The first layer is a convolutional layer with 300 filters and a kernel size of $(5,5)$, where the kernel is padded so that the full kernel lies in the input matrix. The next layer is a pooling layer with kernel size $(3,3)$, where the kernel is padded to preserve the shape of the input. This is followed by a second convolutional layer with 150 filters and a kernel size of $(5,5)$. This is followed by a pooling layer with the same properties as before. Next, the outputs are flattened into dense layers of shape $[384,192,10]$. In Fig.\ref{fig:LeNet} we can see a graphical representation of the network structure.
All layers except the pooling layer, the flattening layer, and the last layer have a ReLU activation function. The last layer has a softmax activation function.
Layer weights are initialized with Glorot uniform. Biases are set to zero.
The network is trained on the CIFAR-10 dataset for 100 epochs with a learning rate schedule $\eta(t)=0.01\times0.95^{t},$ where time steps $t$ are in epochs, the final learning rate is $\eta(T)\approx5.9\times10^{-5}$. The loss used is the cross entropy and the update rule is the SGD with a batch size of 32. We use a momentum of $\beta=0.9$.
All dense layers are regularized with a weight decay of $\lambda=0.0001$.
After training, the network reaches a full training accuracy of $100\%$ and a test accuracy of $\approx 74.9\%,$ depending on the seed. In its original form, there is an additional layer, the local response normalization layer. This layer is not used here because it causes problems in newer versions of TensorFlow and is considered outdated and more or less unimportant in terms of generalization performance \cite{simonyan2014very}.

\subsubsection{ResNet20}
\label{sec:resnet}
ResNet \cite{he2016deep} is a convolutional network that additionally uses batch normalization and, most importantly, residual layers.
The first layer is a convolutional layer with 16 filters and a kernel size of $(3,3)$, where the kernel is padded to preserve the shape of the input. The layer uses batch normalization followed by the ReLU activation function. After this layer, the residual block comes into play. The residual block, where the residual layers lie in, consists of six convolutional layers with $16\times2^{n_\text{block}-1}$ filters, where $n_\text{block}$ is the block counter, a number that starts with $1$ and increase with each following additional block, and a kernel size of $(3,3)$. After each convolutional layer, batch normalization and the ReLU activation function are applied. After every second layer, a residual layer is added between the batch normalization and the activation function, connecting the output of the second layer to the input of the first layer. For our ResNet, we use three of these residual blocks. This is followed by an average pooling layer, which averages the spatial indices coming out of the residual block. Finally, the output is flattened with a softmax activation function to match the output dense layer of size 10.
Layer weights are initialized with the Glorot normal. Biases are set to zero.
The network is trained on the CIFAR-10 dataset for 160 epochs with a learning rate schedule of 
\begin{equation}
\eta(t)=
    \begin{cases}
    10^{-3}&t<80\\
    10^{-4} &120>t\geq80\\
    10^{-5}&t\geq120\, ,
    \end{cases}
\end{equation}
where the time steps $t$ are in epochs. The loss used is the cross entropy and the update rule is the SGD with a batch size of 128. We use a momentum of $\beta=0.9$.
All layers are regularized with a weight decay of $\lambda=0.0001$.
After training, the network reaches a full training accuracy of $100\%$ and a test accuracy of $\approx 68.1\%,$ depending on the seed. At first glance, it looks very similar to LeNet in terms of test accuracy, but it can massively outperform LeNet when using data augmentation. This is the creation of new images by cropping and rotating the original images in the dataset so that they still match the label class. The reason why data augmentation is not used here is that it would lead to the question of what is the total training loss and how to derive the Hessian matrix from it.

\section{Algorithms for Deep Neural Network Analysis}
The code used for this thesis is written entirely in Python and can be found on \url{https://github.com/RosenowGroup/Hessian-eigenvectors-PCA-DNN-weights}. 
The modules used, and their most important functions are explained below.
\subsection*{Deep Neural Network Framework}
The Python module TensorFlow \cite{tensorflow2015-whitepaper} is used to train and analyze the networks. This framework provides almost all important functions for building and testing networks. In addition, it comes with its own type of tensors that can be used to build special TensorFlow functions that are much faster than Python functions because they allow parallelization and use of the graphics processing unit (GPU). Another very useful tool in TensorFlow is auto-differentiation. TensorFlow keeps track of all the operations used to compute e.g. the loss and allows a numerically optimal, precise evaluation of the gradient through predefined exact derivatives. We can expect all results to be independent of the framework. Deviations can be caused by different random seeds, e.g. during initialization, or by numerical instabilities during training.
\subsection*{Computation of the Hessian matrix}
The Hessian matrix must be computed as a function of all training images or a sufficiently large subset. We need to extract the second derivative of the loss. To do this, we can use the auto-differentiation on the first derivative. In Lst.\ref{lst:hessian} we can see how to compute a Hessian vector product (hvp) $\bm{H}\bm{v}$. 
\verb|model| is the model function that returns the softmax predictions of an input. The \verb|loss_object| computes the loss function from the network's predictions and the actual true labels. \verb|tf.GradientTape| tells TensorFlow to record all the operations it performs, from which we can take the gradient \verb|g.gradient| from an output to an input. \verb|output_gradients=vector| tells the gradient tape to compute the vector product of the gradient matrix and the vector. To compute the Hessian matrix, we can use the Hessian vector product of the standard basis to reconstruct its rows. While TensorFlow offers the possibility to compute the Hessian directly, this requires more memory and can lead to bugs in older TensorFlow versions (we use TensorFlow 2.4). Another method to extract the eigenvalues and eigenvectors is to use the Lanczos algorithm \cite{lanczos1950iteration}. Here we start with a random vector from which we compute the hvp, extract a new vector, which is then used for the next hvp. Running this algorithm $n$ times iteratively yields a tridiagonal matrix with $n-1$ eigenvalues that approximate the largest eigenvalues in magnitude. In addition, the eigenvectors can also be extracted. The original algorithm leads to numerical instability, which can be avoided by several methods. One is to use \verb|science.sparse.linalg.eigsh| from the SciPy \cite{2020SciPy-NMeth} module, which is accurate but very slow since it requires computing twice the number of hvp of the desired eigenvectors. Another faster but less accurate implementation can be found in the GitHub \verb|old| folder of the project for this thesis linked earlier.
\begin{lstlisting}[float,language=Python,label=lst:hessian,caption={Computation of the Hessian vector product in TensorFlow. Here we record the gradient of the gradient of the loss (\lstinline|loss_object|) of the network output (\lstinline|model(images)|) in direction of a given vector (\lstinline|vector|).}]
import tensorflow as tf
with tf.GradientTape() as g:
    with tf.GradientTape() as gg:
        predictions = model(images)
        loss = loss_object(labels, predictions)
    gradient = gg.gradient(loss, weights)
hvp=g.gradient(gradient, weights, output_gradients=vector)
\end{lstlisting}
\subsection*{Computation of tensor objects}
When we measure the weights of the networks, we store them as NumPy arrays. NumPy \cite{harris2020array} is a module that contains many different operations that can be performed on these arrays. These arrays are tensors of arbitrary shape. To compute the covariance matrix, we can simply use \verb+numpy.cov+. To get the principal components, we can then use \verb+numpy.linalg.eigh+ or \verb+tensorflow.linalg.eigh+ to compute the eigenvalues and eigenvectors of a hermitian or symmetric matrix. The TensorFlow function can use the GPU if the memory space is large enough, and it is also more stable when handling larger matrices, but does not return errors if there are discrepancies. Similarly, for SVD, there is a function \verb|linalg.svd| in both modules to compute the singular values and their singular vectors.
\chapter{Analysis of the Dynamic Weight Matrix}
\section{Comparison of Singular Values and Principal Components}
\begin{figure}[t]
	\centering
	\includegraphics[width=0.9\textwidth,angle=0]{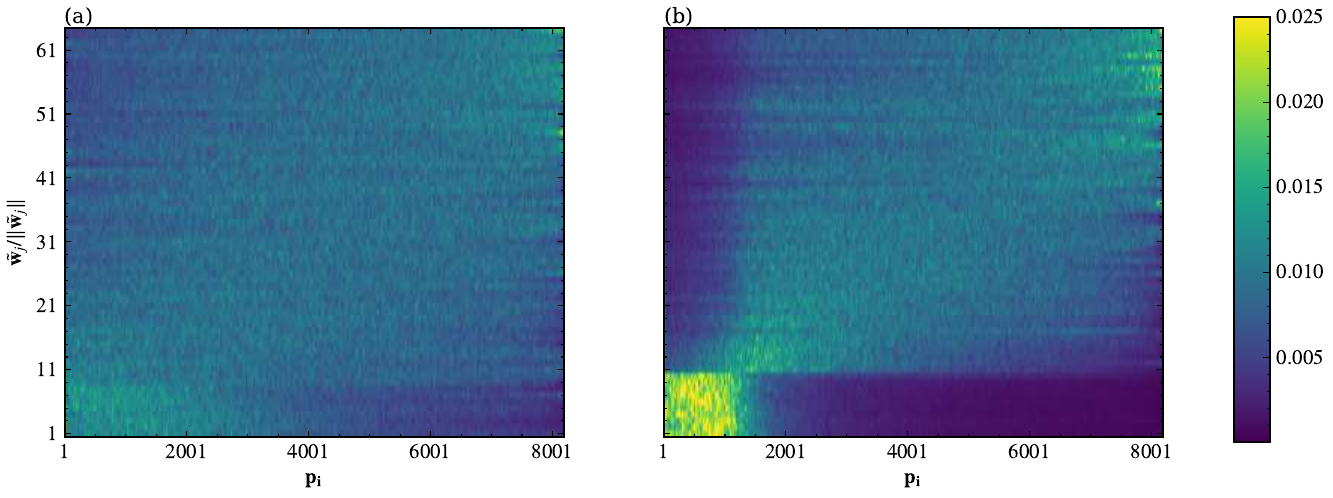}
	\caption{Scalar product of singular matrices and principal components of the $(128,64)$ layer. Measured weights for 10 epochs post-training. (a) Unregularized for all layers. (b) All layers regularized with weight decay of $\lambda=0.0005$.}
	\label{fig:pc_svd}
\end{figure}
\begin{figure}[t]
	\centering
	\includegraphics[width=0.9\textwidth,angle=0]{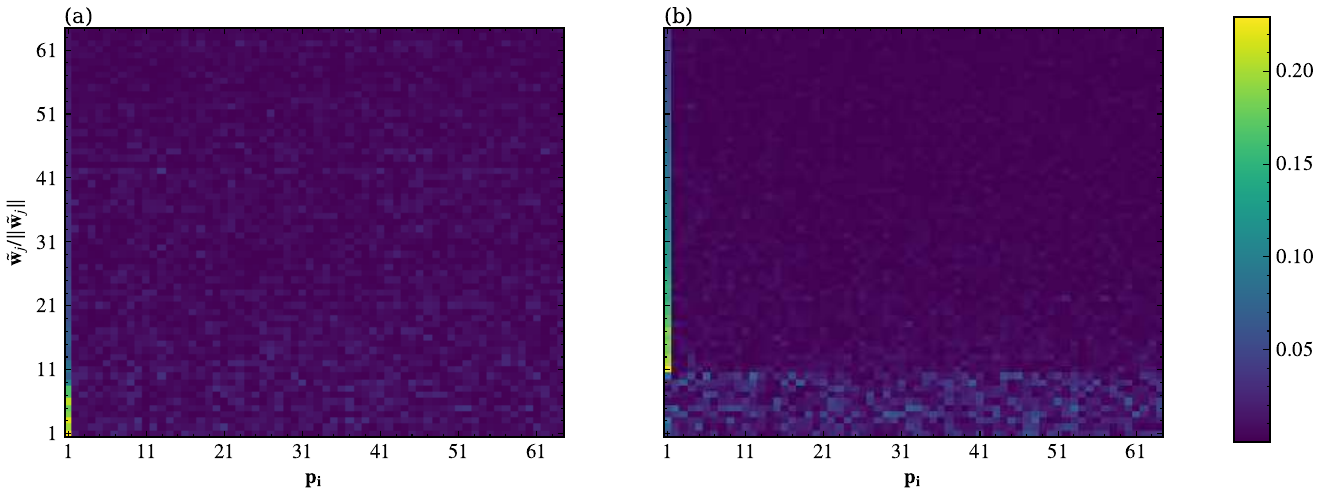}
	\caption{Scalar product of singular matrices and principal components of the $(128,64)$ layer from the MLP 256 network, zoomed in to the largest 64 principal components. Weights measured for 10 epochs post-training. (a) Unregularized for all layers. (b) All layers regularized with weight decay of $\lambda=0.0005$.}
	\label{fig:pc_svd_zoom}
\end{figure}
Our examination begins with a thorough exploration of the relationship between singular values and principal components of the network weights. Although the dimensions of the left and right singular vectors differ from those of the originating weight matrix, a transformation involving matrices $\bm{\tilde{W}}_i$ of single singular values brings these vectors together:
\begin{equation}
	\bm{\tilde{W}}_i=\bm{U}\bm{\tilde{\Sigma}}_i\bm{V}^T, \quad \bm{\tilde{\Sigma}}_i=\text{diag}(0,...,\nu_i,0,...,0)\, .
\end{equation}
Assembling these matrices, each comprising a left and a right singular vector, culminates in the reconstruction of the original weight matrix. The flattening of $\bm{\tilde{W}}_i$ into a vector $\bm{\tilde{w}}_i$ facilitates the analysis of scalar products with the principal components $\bm{p}_i$ using the equation:
\begin{equation}
	\label{eq:sv_vec}
	S_{ij}=\frac{|\bm{p}_i\cdot\bm{\tilde{w}}_j|}{\|\bm{\tilde{w}}_j\|}\, .
\end{equation}
Note that we are considering the absolute, since the choice of sign of the principal components is arbitrary. Unless otherwise noted, all principal component and singular value computations are performed after training. The visualization in Fig.\ref{fig:pc_svd} unveils intriguing insights into the interaction between principal components and singular values. Notably, principal components show an interesting correlation with singular values. Remarkably, the principal components with the lowest indices have the largest product with singular values of the lowest indices. Here, the variances of the principal components and singular values underscores that the principal components with the highest variance shares a strong connection with the singular matrices of singular values situated outside the RMT bulk. At the same time, principal components with lower variances entail smaller product values relative to the former, yet display larger product values relative to singular values residing within the RMT bulk. Given the undeniable significance of singular values outside the RMT bulk as seen in Sec.\ref{sec:RMT}, we infer that the initial principal components encapsulate critical information due to their substantial scalar product. In the context of weight decay, the alignment of the largest singular values with the largest principal components underscores the presence of a distinct boundary between singular values from within and outside the RMT bulk, as depicted by their scalar product with principal components.\\
Zooming in to focus on the largest principal components, Fig.\ref{fig:pc_svd_zoom} exposes that the scalar product of the first principal component surpasses that of the subsequent components by a significant margin. This phenomenon, not apparent in the previous figure due to color resolution, showcases an ordering of scalar products across the first principal component. Remarkably, the scalar product decreases as we examine singular matrices of smaller singular values. For weight decay, the first principal component's scalar product is primarily attributed to the singular matrices within the RMT bulk, as opposed to the training without weight decay, where this occurs predominantly outside the RMT bulk. This phenomenon could be attributed to the comparatively small change in the outlying singular values induced by weight decay, juxtaposed with the substantial decrease of singular values in the bulk necessitated by network optimization, since outlying singular values will stagnate in size.
\section{Principal Components' Influence on Network Performance}
\begin{figure}[t]
	\centering
	\includegraphics[width=0.9\textwidth,angle=0]{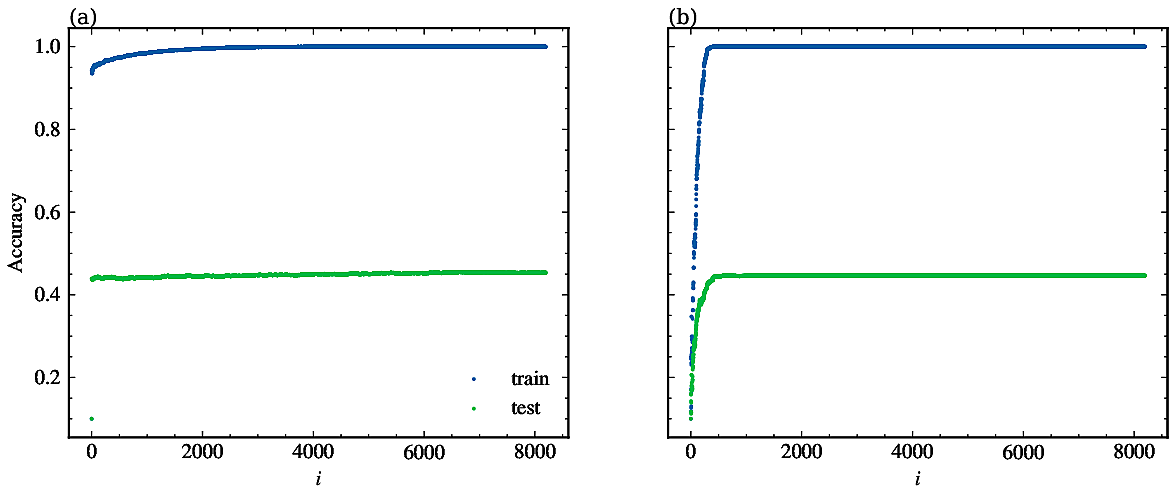}
	\caption{Accuracy of added principal components of the $(128,64)$ layer of the MLP 256 network. (a) All layers are unregularized. (b) All layers are regularized with weight decay $\lambda=0.0005$. The entry $i=0$ corresponds to a weight matrix of zeros.}
	\label{fig:pc_acc}
\end{figure}
Shifting focus, we investigate the influence of principal components on network performance. We can introduce an additive weight vector based on principal components. This can be described as the follows:
\begin{equation}
	\bm{w}_{i,\text{add}}=\sum^i_{j=1}\theta_j(T)\bm{p}_j,\ \theta_j(T):=\bm{w}(T)\cdot\bm{p}_j\, ,
\end{equation}
where $T$ represents the time of training completion.
Fig.\ref{fig:pc_acc} showcases the impact of principal component addition on network accuracy. Evidently, the unregularized network harbors substantial information within the first principal component, emphasizing the dominance of the first principal components. Conversely, the network employing weight decay necessitates the incorporation of approximately the first 500 largest principal components to achieve peak accuracy. This divergence could be attributed to the fact that the first principal component, as seen in Fig.\ref{fig:pc_svd_zoom} with weight decay, no longer covers the outlying singular values primarily.
Furthermore, we avoid the training on a subset of trainable variables during measurement to avoid undesirable outcomes. If a single layer is measured and the layers between that layer and the output are not updated, the high scalar product of the first principal component with the weights will disappear, and the loss will almost stay constant for further training. To avoid this behavior, which is not a practical training dynamic, we always update all layers.
\section{Analysis of the Drift Mode}
\label{sec:drift}
\begin{figure}[t]
	\centering
	\includegraphics[width=0.55\textwidth,angle=0]{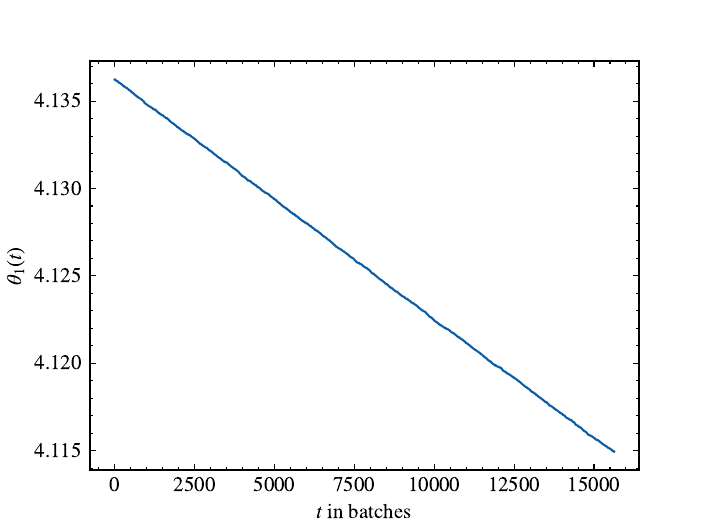}
	\caption{Weight development in direction of the first principal component of the $(128,64)$ layer of the regularized MLP 256 network.}
	\label{fig:theta}
\end{figure}
\begin{figure}[t]
	\centering
	\includegraphics[width=0.5\textwidth,angle=0]{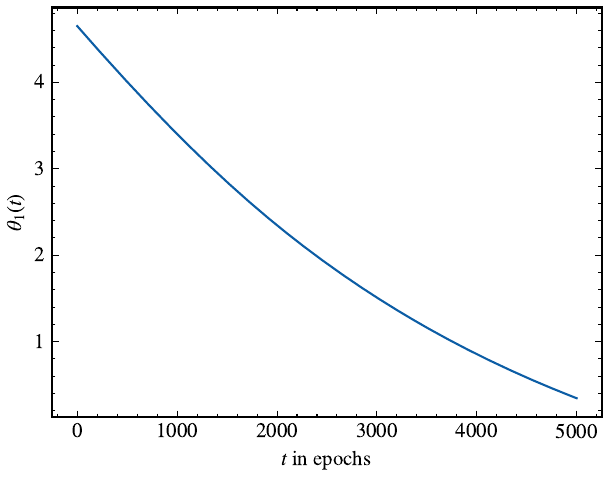}
	\caption{Weight development in direction of the first principal component of the $(128,64)$ layer of the regularized MLP 256 network. Time s are measured in epochs.}
	\label{fig:theta_long}
\end{figure}
\begin{figure}[t]
	\centering
	\includegraphics[width=0.5\textwidth,angle=0]{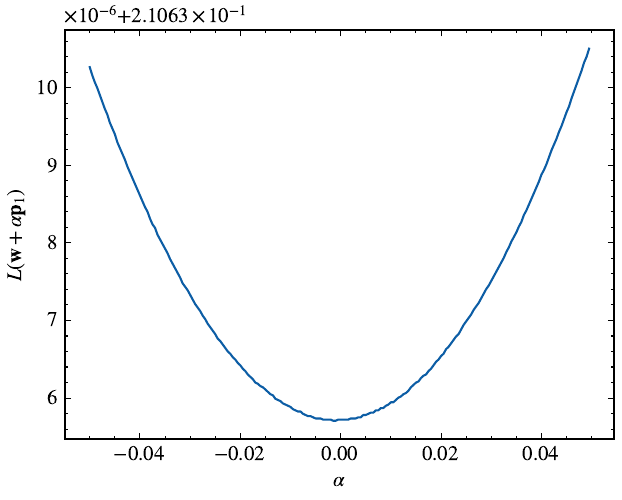}
	\caption{The loss landscape in direction of the drift mode of the $(128,64)$ layer of the regularized MLP 256 network. The network is very close to the minimum in direction of the drift mode. Each update step only moves the network very little further to the minimum.}
	\label{fig:drift_pot}
\end{figure}

In the direction of the first principal component, a clear linear behavior emerges, as depicted in Fig.\ref{fig:theta}. This behavior is well-captured by:
\begin{equation}
	\theta_1(t)=\bm{w}(t)\cdot\bm{p}_1\approx a(t-t_0)+b,\quad t \geq t_0\, ,
\end{equation}
where $t_0$ marks the commencement of our measurement during the exploration phase, $a$ is the slope and $b=\bm{w}(t_0)\cdot\bm{p}_1$. Notably, all other principal components deviate from linearity and exhibit a random walk pattern, with their effects on the network's weights remaining neglectable.\\
Moreover, the dynamics of unregularized networks reveal that $|\theta_1(t)|$ experiences growth over time. In contrast, networks featuring weight decay exhibit the shrinkage of both $|\theta_1(t)|$ and $\|\bm{w}\|$.
Considering the evolution of $\|\bm{w}(t)\|$, we discern an almost linear function approximately as for $|\theta_1(t)|$.
We refer to the movement of the network as the "drift mode" for its conspicuous linear evolution \cite{doi:10.1073/pnas.2015617118}.
This observation paves the way for the approximation
\begin{equation}
	\bm{w}(t)\approx \bm{a}(t-t_0)+\bm{w}(t_0)\, ,
\end{equation}
where $\bm{a}=a\bm{p}_1$. This result stems from the fact that only the drift mode is responsible for persistent changes within the network.\\
The quantification of the drift mode contribution leads us to derive its variance to gain deeper insights. Mathematically, the variance $\sigma^2$ can be expressed as:
\begin{equation}
	\begin{split}
		\sigma^2_1&=\langle\theta_1(t)^2\rangle-\langle\theta_1(t)\rangle^2 \\
		&=\frac{1}T\sum^{T+t_0}_{t=t_0}\left(a^2(t-t_0)^2+b^2+2a(t-t_0)b\right)-\left(\frac{1}T\sum^{T+t_0}_{t=t_0}(at+b)\right)^2 \\
		&=\frac{a^2}{12}(T^2-1)\, ,
	\end{split}
\end{equation}
where $T$ signifies the number of measured time s.
The application of Faulhaber's formula facilitates the solution of the sums, yielding a clear representation of the variance as a function of $a$, $T$, and the network's configuration. Since each update step points approximately in the same direction, we can connect the slope to the learning rate by using the SGD:
\begin{equation} 
	a\approx\|\frac{\bm{w}(t)-\bm{w}(t_0)}{t-t_0}\|\propto \eta\, .
\end{equation} 
As observed previously, when we extend the length of the observation interval or increase the size of the learning rate, the eigenvector of the largest eigenvalue of the covariance matrix approximates the drift mode, but only if it has a larger variance than those of the largest noise. This intriguing relationship can be validated by training the network up to the exploration phase using a high learning rate, effectively mitigating computational time constraints for low learning rates. The observed relationship, $\sigma^2\propto\eta^2T^2$, holds for appropriate learning rates and measurement time s. Furthermore, for extensive epochs, here for this comparatively small network $\approx 600$ epochs, the feasibility of measuring every batch becomes constrained due to memory limitations. However, given the drift mode's computationally efficient characterization via a linear fit, it suffices to measure weights at more extended intervals, such as epochs. Remarkably, as demonstrated by the MLP 256 network in Fig.\ref{fig:theta_long}, the drift mode ceases to exhibit a linear trend beyond approximately 500 epochs. This indicates the presence of higher-order terms contributing to the observed behavior.\\
To get a better understanding on this evolving behavior, a detailed analysis of the loss landscape within the drift mode's direction is crucial. The depiction in Fig.\ref{fig:drift_pot} illustrates the presence of a potential well aligned with the drift mode's direction. Notably, this characteristic potential well corresponds to the presence of similar potential wells aligned with the Hessian eigenvectors, as described in Eq.\eqref{eq:loss_hess_quad}. Importantly, this phenomenon is qualitatively observed across our used network architectures, limiting the explanation, that its causality being rooted solely in regularization effects.\\
The drift mode's dynamic behavior, marked by adjustments toward the loss minimum, offers insights into network training dynamics. Nonetheless, the minuscule magnitude of these adjustments implies that these insights may not be of great importance within the context of training for optimal performance.

\section{A Model for the Drift Mode}
We begin by considering the quadratic loss approximation:
\begin{equation}
	L(\bm{w})\approx L_{\text{min}}+(\bm{w}-\bm{\mu})^T\bm{\frac{H}{2}}(\bm{w}-\bm{\mu})\, ,
\end{equation}
where $L_\text{min}$ is the minimum of the loss, $\bm{\mu}$ is the point of the minimum in weight space and $\bm{H}$ is the Hessian matrix.
We neglect momentum and include weight decay parametrized by its strength $\lambda$. The average update step of the Stochastic Gradient Descent is given by:
\begin{equation}
	\langle\frac{d\bm{w}(t)}{dt}\rangle_\mathcal{B}=-\frac{\eta}{S}(\bm{H}+2\lambda)\bm{w}+\eta \bm{H}\bm{\mu}\, ,
\end{equation}
where $\langle\cdot\rangle_\mathcal{B}$ is the average over all batches and $S$ the batch size.
Let us assume that during the exploration phase, each update step is similar to the average update:
\begin{equation}
	\frac{d\bm{w}(t)}{dt}\approx\langle\frac{d\bm{w}(t)}{dt}\rangle_\mathcal{B}\, .
\end{equation}
This makes the differential equation linear and of first order, allowing us to solve it explicitly:
\begin{equation}
	\label{eq:exp_decay}
	\hat{w}_i(t)=\left(\hat{w}_i(0)-\frac{\hat{\mu}_i}{1+2\lambda/h_{i}}\right)e^{-\eta (h_{i}+\lambda)t}+\frac{\hat{\mu}_i}{1+2\lambda/h_{i}}\, ,
\end{equation}
where $\hat{w}_i(t)=\bm{w}(t)\cdot\bm{h}_i$, $\hat{\mu}_i=\bm{\mu}\cdot\bm{h}_i$ and $h_i$ are the Hessian eigenvalues their corresponding eigenvectors $\bm{h}_i$. When employing weight decay, the network's minimum is shifted to lower values. The drift mode can be decomposed using Hessian eigenvectors, since those form an eigenbasis in the weight space. The first-order approximation of the drift mode can be related to the observed drift mode in Sec.\ref{sec:drift}. For longer measurements, an exponential decay is expected, as observed in Fig.\ref{fig:theta_long}. The variance is computed using:
\begin{equation}
	\begin{split}
		\sigma^2_i
		&=\langle \hat{w}_i(t)^2\rangle-\langle \hat{w}_i(t)\rangle^2
		\\&=\frac{1}{T}\sum^T_{t=0}\left(\left[\hat{w}_i(0)-b_i\right]^2e^{-2\tilde{\lambda}_it}+b_i^2+2\left[\hat{w}_i(0)-b_i\right]b_ie^{-\tilde{\lambda}_it}\right)
		\\&\quad -\left(\frac{1}{T}\sum^T_{t=0}\left[(\hat{w}_i(0)-b_i)e^{-\tilde{\lambda}_it}+b_i\right]\right)^2
		\\&=\frac{1}{T}\sum^T_{t=0}(\hat{w}_i(0)-b_i)^2e^{-2\tilde{\lambda}_it}-\frac{1}{T^2}\left(\sum^T_{t=0}(\hat{w}_i(0)-b_i)e^{-\tilde{\lambda}_it}\right)^2
		\\&=\frac{(\hat{w}_i(0)-b_i)^2}{T}\left(\frac{1-e^{-2\tilde{\lambda}_i(T+1)}}{1-e^{-2\tilde{\lambda}_i}}-\frac{1}{T}\left(\frac{1-e^{-\tilde{\lambda}_i(T+1)}}{1-e^{-\tilde{\lambda}_i}}\right)^2\right)\, ,
	\end{split}
\end{equation}
where $\tilde{\lambda}_i:=\eta (h_{i}+2\lambda)$ and $b_i:=\frac{\hat{\mu}_i}{1+2\lambda/h_{i}}$. The geometric sum formula is used to solve the sums. Notably, the variance of the drift converges to zero as time increases, indicating that it does not increase indefinitely. In this model, the drift mode is a sum of the Hessian eigenvectors. If we decompose the drift as $\bm{p}_1=\sum_id_i\bm{h}_i,$ with $d_i:=\bm{h}_i\cdot\bm{p}_1,$ then the variance of the drift mode is given by:
\begin{equation}
	\sigma^2_d=\sum_id_i^2\sigma^2_i\, .
\end{equation}
The coefficients $d_i$ with $\sum_id^2_i=1$ are chosen such that $\sigma^2_d$ maximizes when the drift mode aligns with the first principal component.

\section{Loss Scaling}
\label{sec:loss_scaling}
\begin{figure}[t]
	\centering
	\includegraphics[width=0.9\textwidth,angle=0]{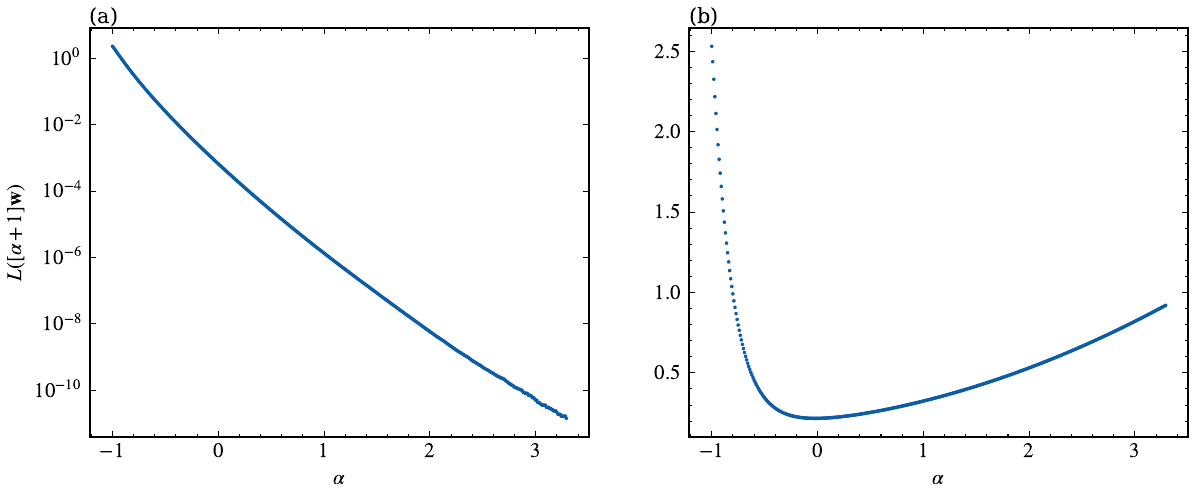}
	\caption{Loss scaling of the (128,64) layer of the MLP 256 network. (a) All layers are unregularized. (b) All layers are regularized with weight decay $\lambda=0.0005$.}
	\label{fig:loss_scale}
\end{figure}

Consider an unregularized network employing softmax that achieves a training accuracy of $100\%$. If we increase the weights' size by a factor of $\alpha$, it will enhance confidence in the probability distribution and consequently reduce the loss, without altering the training accuracy. Assuming that the logits of the correct labels significantly exceed the others, i.e., $z_j \gg z_k, \forall k\neq j$, we can demonstrate that:
\begin{equation}
	\begin{split}
		L(\alpha\bm{z})
		&=-\ln\frac{e^{\alpha z_j}}{\sum_ie^{\alpha z_i}} \\&=-\alpha z_j+\alpha z_j+\ln(1+\sum_{i\neq j}e^{\alpha(z_i-z_j)}) \\&=\sum_{i\neq j}e^{\alpha(z_i-z_j)}+O\left(\left[\sum_{i\neq j}e^{\alpha(z_i-z_j)}\right]^2\right) \\&\propto e^{-\alpha z_j}\, .
	\end{split}
\end{equation}
Using ReLU and scaling the weights layer by layer, we expect $z_i\propto\alpha\sum_{j} a_j w_j,\ \forall i,$ where $a_j$ are coefficients dependent on the input. Fig.\ref{fig:loss_scale} illustrates that this assumption agrees with experimental results for the unregularized network, considering that scaling the weights proportionally changes the logits. Interestingly, while numerically increasing the weights can drive the loss to zero in this direction, the network tends to move in the direction of the drift mode instead. Both directions point in a similar direction, considering their large product $\approx 0.7$ here, but they are not completely parallel. For the regularized network, we observe that $L\propto\alpha^2$ when scaling the network weights further. Therefore, scaling a network with weight decay can not reduce the loss further.

\chapter{Analysis of the Hessian Matrix}
In this chapter, we delve deeper into the eigenvectors of the Hessian matrix.
Let $\bm{H}\in \mathbb{R}^{n\times n}$ be the Hessian matrix with eigenvectors $\bm{h}_i$ and corresponding eigenvalues $h_i$:
\begin{equation}
	\bm{H}\bm{h}_i=h_i\bm{h}_i,\ h_i\geq h_{i+1}\, .
\end{equation}
Because the Hessian matrix is symmetric, its eigenvalues are real, and its eigenvectors form an orthonormal basis.\\
Starting from the quadratic loss approximation in Eq.\eqref{eq:loss_hess_quad}, we can interpret positive eigenvalues as measures of the curvature of minima along the corresponding eigenvector direction. Hence, eigenvectors corresponding to the largest eigenvalues are the crucial directions for loss minimization \cite{doi:10.1073/pnas.2015617118}. These eigenvectors define regions in which the network must be closer to the minimum $\bm{\mu}\cdot \bm{h}_i$ than in directions with smaller positive eigenvalues.

\section{Hessian Eigenvectors and PCA}
\begin{figure}[t]
	\centering
	\includegraphics[width=0.9\textwidth,angle=0]{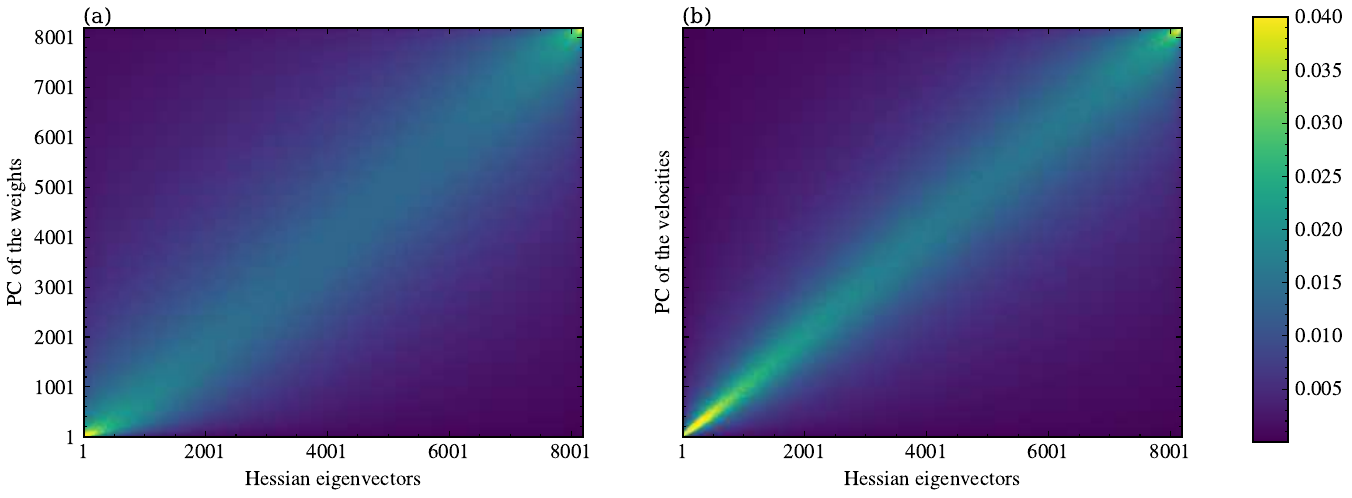}
	\caption{Scalar product of the Hessian eigenvectors with the principal components of (a) weights and (b) velocities of the $(128,64)$ layer of the unregularized MLP 256 network.}
	\label{fig:phc}
\end{figure}

By comparing the absolute scalar product of the principal components of weights and velocities $\bm{v}(t+1)=\bm{w}(t+1)-\bm{w}(t)$ with the Hessian eigenvectors, as depicted in Fig.\ref{fig:phc}, we observe that the eigenvectors for the largest and smallest eigenvalues are quite similar for all bases. For intermediate eigenvalues, the similarity is diminished, but they still maintain some degree of diagonalization. Further, we observe that the velocity covariance matrix exhibits a sharper diagonal in the Hessian eigenbasis than for the eigenvectors of the weight covariance matrix. This indicates that the velocity eigenbasis is a more suitable choice for approximating the Hessian eigenbasis with a covariance matrix, potentially reducing computational time \cite{doi:10.1073/pnas.2015617118}. Notably, measuring velocities instead of weights incurs no additional computational cost, as both are calculated for each update step.

\section{Eigenvectors and the Weight Product}
\label{sec:w_prod}
\begin{figure}[t]
	\centering
	\includegraphics[width=0.9\textwidth,angle=0]{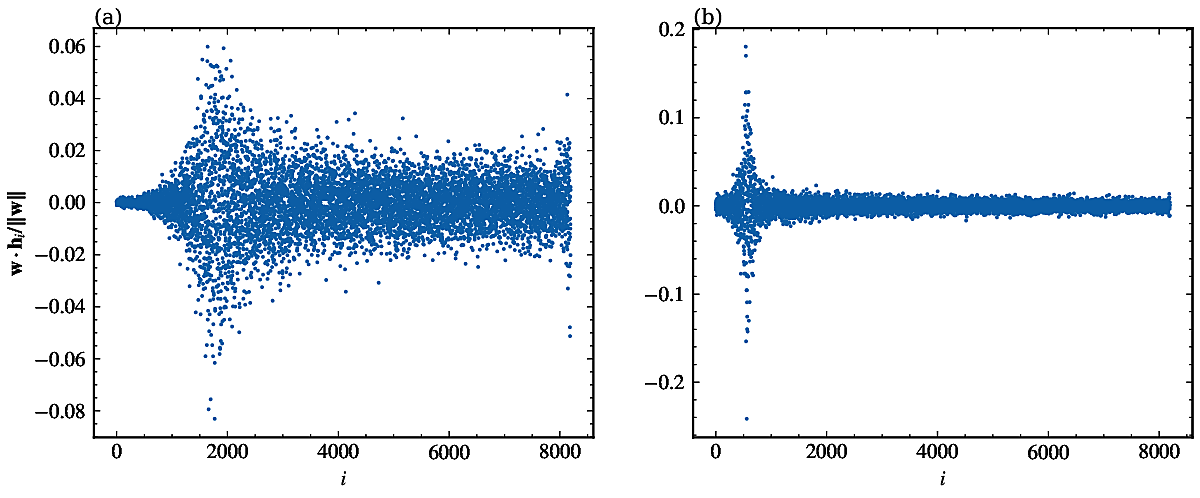}
	\caption{Weight product of the $(128,64)$ layer of the regularized MLP 256 network. (a) All layers are unregularized. (b) All layers are regularized with weight decay $\lambda=0.0005$.}
	\label{fig:hvw}
\end{figure}
\begin{figure}[t]
	\centering
	\includegraphics[width=0.9\textwidth,angle=0]{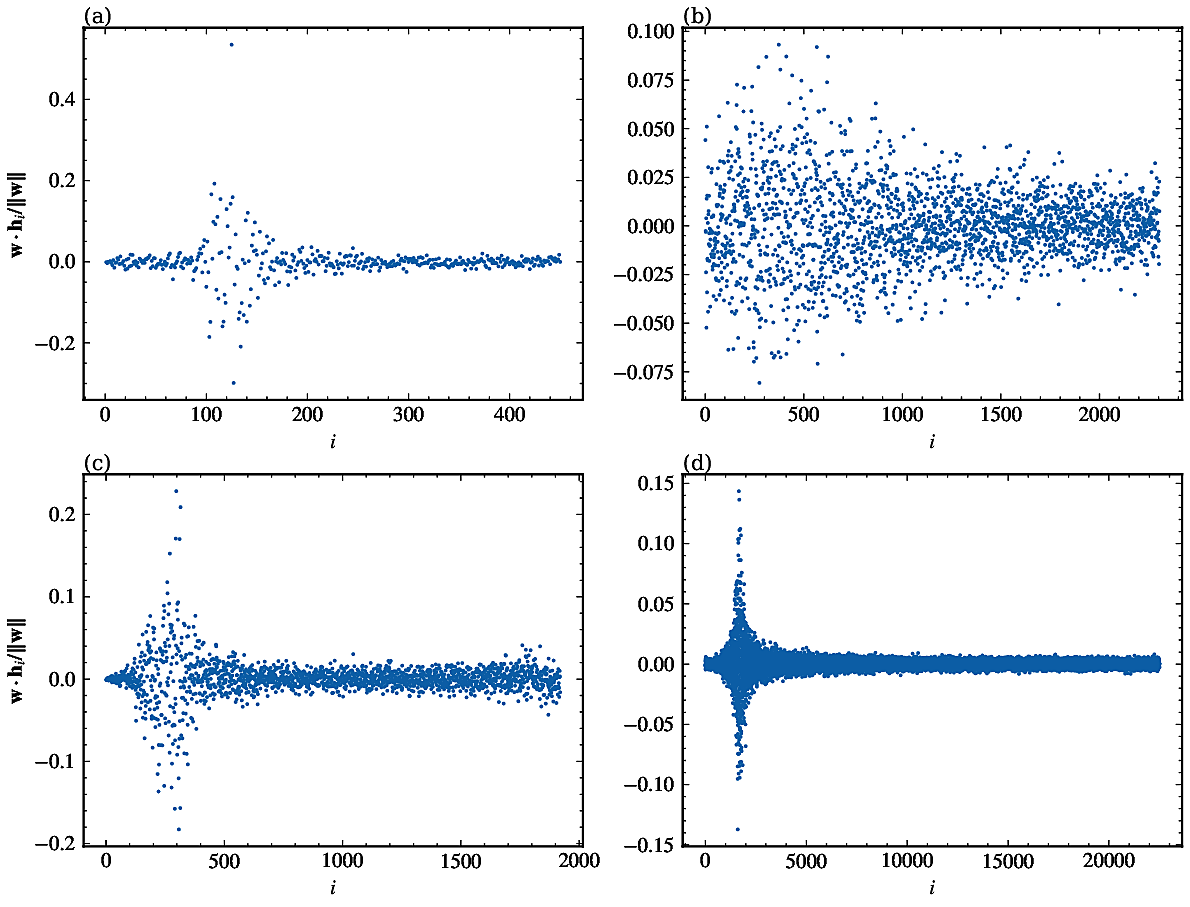}
	\caption{Weight product for various networks and layers: (a) First convolutional layer of LeNet. (b) Second convolutional layer of ResNet. (c) Last layer of miniAlexNet. (d) First convolutional layer of miniAlexNet.}
	\label{fig:hvwm}
\end{figure}
\begin{figure}[t]
	\centering
	\includegraphics[width=0.9\textwidth,angle=0]{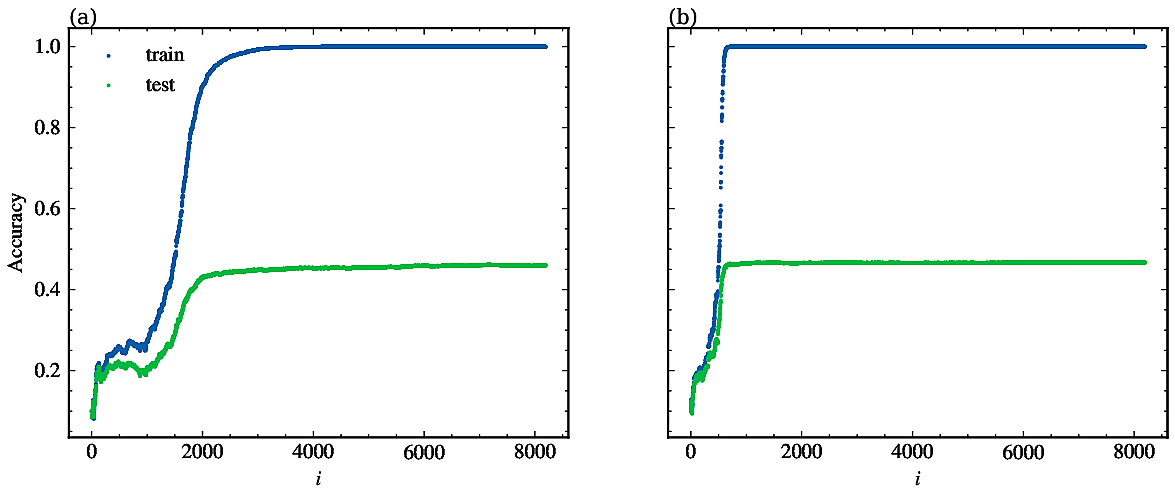}
	\caption{Accuracy of added Hessian eigenvectors of the $(128,64)$ layer of the MLP 256 network. (a) All layers are unregularized. (b) All layers are regularized with weight decay $\lambda=0.0005$.}
	\label{fig:hacc}
\end{figure}

Consider the scalar product of the Hessian eigenvectors with the network weights:
\begin{equation}
	\frac{\bm{h}_i\cdot\bm{w}(T)}{\|\bm{w}(T)\|}\, ,
\end{equation}
referred to as the weight product. In Fig.\ref{fig:hvw}, it is evident that without weight decay, eigenvectors associated with larger eigenvalues have a smaller weight product with the network, compared to randomly distributed eigenvectors associated with smaller eigenvalues. When weight decay is employed, these values become comparable in magnitude. For both scenarios, the largest scalar products are found between the 500th and 2000th eigenvectors in proximity to each other. For large enough index ranges $\gtrsim 500$, the weight product distribution appears Gaussian in local regions with varying variances when testing those on the distribution. This behavior is anticipated in all networks. An explanation could be that we initialize layers uniformly or normally. The central limit theorem establishes that the sum of uniformly drawn numbers converges to a Gaussian distribution. Assuming that the weights are mainly random, which is underlined by the fact that the network only trained in a small part of the Hessian eigenvectors, this can explain why the weight product still follow the distribution.\\
The locations of the largest weight products, especially in unregularized networks, could potentially be due to the loss scaling discussed in Sec.\ref{sec:loss_scaling}. The absence of a potential well in the direction of the entire network, primarily encompassed by large entries, might cause the potential wells of eigenvectors associated with these large entries to not be the steepest. Additionally, symmetry properties of the network could account for the small products observed with the largest eigenvalues. If we presume that eigenvectors corresponding to the largest eigenvalues are translations of labels or rotations, altering entries within a layer could lead to a substantial loss change, potentially misclassifying all labels \cite{guth2023rainbow}.
For other networks and layers, as depicted in Fig.\ref{fig:hvwm}, the precise positioning of the large weight products within the Hessian spectrum remains without a model. Nonetheless, it is clear that layers with more parameters correspond to higher indices where the product is large. \\
Let us now decompose the weights of the layer into the eigenbasis of the Hessian:
\begin{equation}
	\label{eq:adding_acc}
	\bm{w}_{i,\text{add}}=\sum^i_{j=1}\theta_j\bm{h}_j, \ \theta_j:=\bm{w}(T)\cdot \bm{h}_j\, .
\end{equation}
In Fig.\ref{fig:hacc}, it becomes evident that eigenvectors with the largest weight product must be aggregated to achieve optimal network performance. Setting all eigenvectors to zero except those with the largest product does not lead to accuracy degradation. For eigenvectors with small eigenvalues, this is easily explained by arguing that changing $\theta_j$ from a certain small product in magnitude with the weights to zero will not change the loss much because their potential well is very flat. For the eigenvectors of the largest eigenvalues, this may be similar, since their weight product is small in magnitude, but it is not straightforward that because of their steep potential well, even small changes in $\theta_j$ should not change the loss, and thus the accuracy, at all. This will be discussed more in detail for the whole network analysis. For the eigenvectors with a large product, it is plausible that setting $\theta_j$ to zero will change the loss greatly, because $\theta_j$ is large compared to other products and the eigenvalue is in the regime of being larger than most other eigenvalues of the Hessian. Comparing Fig.\ref{fig:hvw} and Fig.\ref{fig:hacc}, we may also conclude that only a fraction of about $\lesssim 20\%$ of the largest Hessian eigenvalues is sufficient to fully describe the network performance. For other networks the behavior is similar. However, the position for the large weight product differs, and with that the fraction of important eigenvalues.
\section{Comparison of Singular Values and Hessian Eigenvectors}
\label{sec:svhess}
\begin{figure}[t]
	\centering
	\includegraphics[width=0.9\textwidth,angle=0]{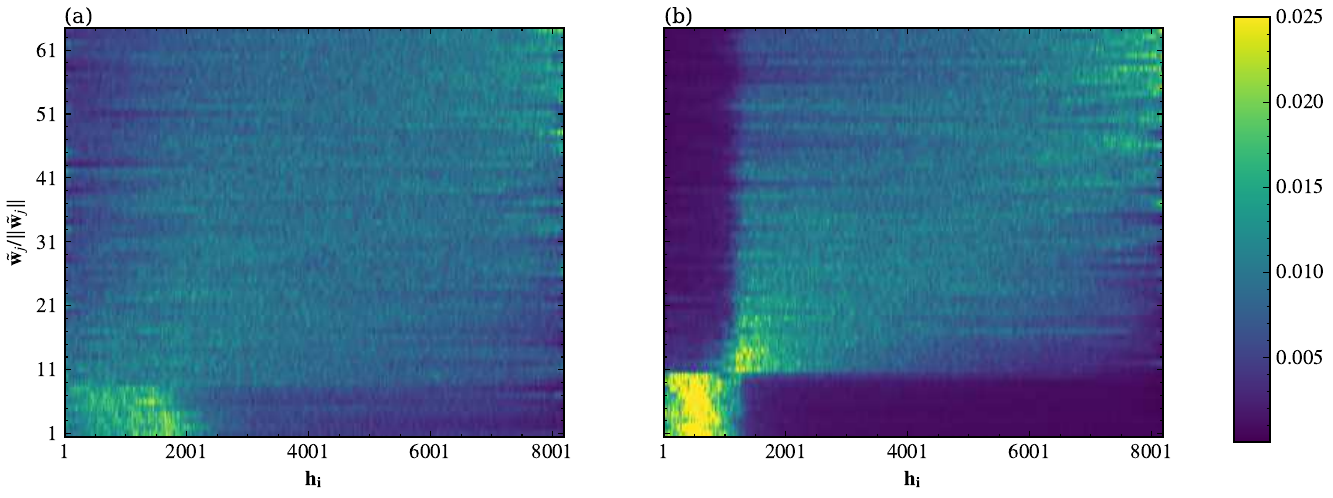}
	\caption{Scalar product of singular matrices and Hessian eigenvectors for the $(128,64)$ layer of the MLP 256 network. (a) All layers are unregularized. (b) All layers are regularized with weight decay $\lambda=0.0005$.}
	\label{fig:hc_svd}
\end{figure}
\begin{figure}[t]
	\centering
	\includegraphics[width=0.5\textwidth,angle=0]{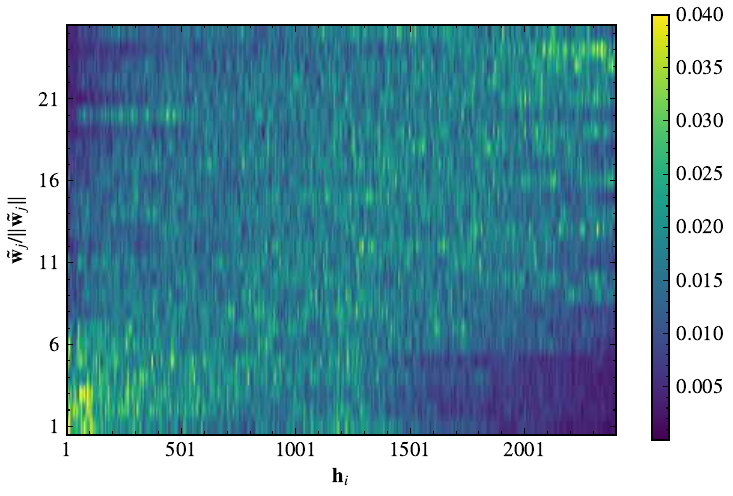}
	\caption{Scalar product of the flattened singular matrices and Hessian eigenvectors for the $(16,(5,5))$ convolutional layer of the LeNet.}
	\label{fig:field_conv}
\end{figure}

Let us explore the relationship between the product of Hessian eigenvectors and singular matrices:
\begin{equation}
	F_{ij}=\frac{|\bm{h}_i\cdot\bm{\tilde{w}}_j|}{\|\bm{\tilde{w}}_j\|}\, .
\end{equation}
As depicted in Fig.\ref{fig:hc_svd}, singular values lying outside the RMT bulk are predominantly covered by Hessian eigenvectors corresponding to the smallest indices. Additionally, singular values within the RMT bulk exhibit larger scalar products with eigenvectors associated with smaller Hessian eigenvalues.\\
In the context of the networks used, we notice that singular values beyond the bulk almost entirely encompass the largest Hessian eigenvalues. This observation elucidates why the largest singular values are adequate to maintain the accuracy demonstrated in \cite{Staats.2022}, as they encapsulate the directions from Fig.\ref{fig:hacc} that are sufficient to achieve full training and test accuracy. A comparison between these findings and those related to principal components (Fig.\ref{fig:pc_svd}) reveals that both the Hessian eigenvector basis and the principal components behave similarly, likely due to their near-diagonal relationship with each other (Fig.\ref{fig:phc}). In scenarios where weight decay is employed, the arrangement of the largest singular values that cover the prominent Hessian eigenvalues becomes more pronounced.
The validation of these results extends to convolutional layers as well. For this purpose, we must transform the four-dimensional tensor into a two-dimensional matrix. Various methods can be employed for this reshaping process. Previous work \cite{Staats.2022, DBLP:journals/corr/abs-1810-01075} has suggested that the specific reshaping technique may not significantly impact the results. Therefore, we opt to reshape the first two dimensions and the last two dimensions together. This approach ensures that we attain the maximum possible number of singular values for the given layer. Remarkably, as illustrated in Fig.\ref{fig:field_conv}, this layer-wise analysis produces qualitatively similar behaviors when the layer has an adequate number of singular values $\gtrsim 20$.

\section{Comparison of the Hessian of Layers and the Full Network}
\label{sec:full_hess}
\begin{figure}[t]
	\centering
	\includegraphics[width=0.5\textwidth,angle=0]{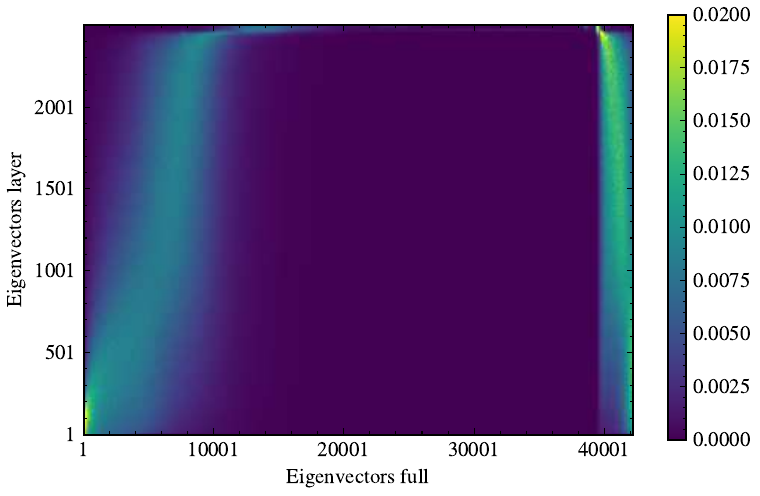}
	\caption{Scalar product of the Hessian eigenvectors of the full MLP 50 network with the Hessian eigenvectors of the $(50,50)$ layer.}
	\label{fig:netw_layer}
\end{figure}
\begin{figure}[t]
	\centering
	\includegraphics[width=0.5\textwidth,angle=0]{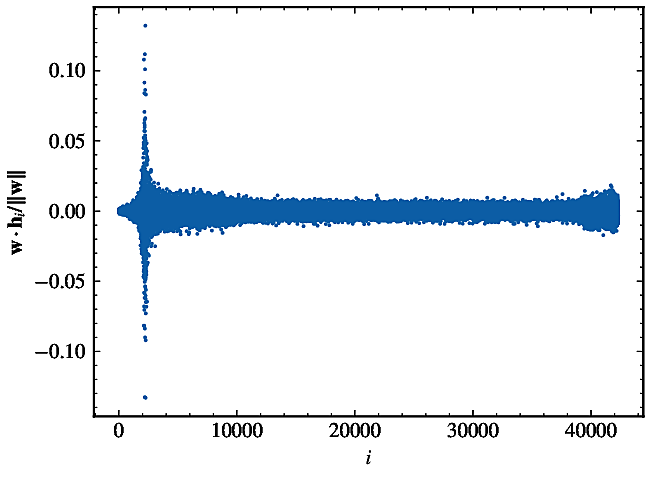}
	\caption{Weight product of the entire MLP 50 network trained on the MNIST dataset. The larger product of the largest indices refer to negative eigenvalues that are of similar size as indices in the range of 8000 to 10000.}
	\label{fig:network_prod_all}
\end{figure}
\begin{figure}[t]
	\centering
	\includegraphics[width=0.9\textwidth,angle=0]{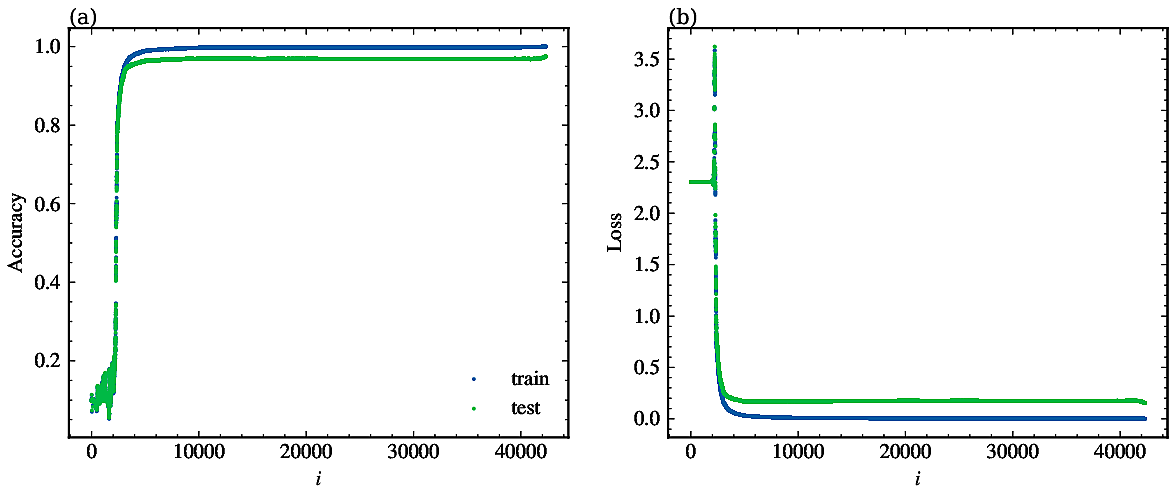}
	\caption{Accuracy of the network with weights of added Hessian eigenvectors of the entire MLP 50 network. The eigenvalues are sorted in decreasing algebraic order. The changes for the largest indices mark negative eigenvalues. (a) Accuracy. (b) Loss.}
	\label{fig:hc_acc_nabs}
\end{figure}
\begin{figure}[t]
	\centering
	\includegraphics[width=0.9\textwidth,angle=0]{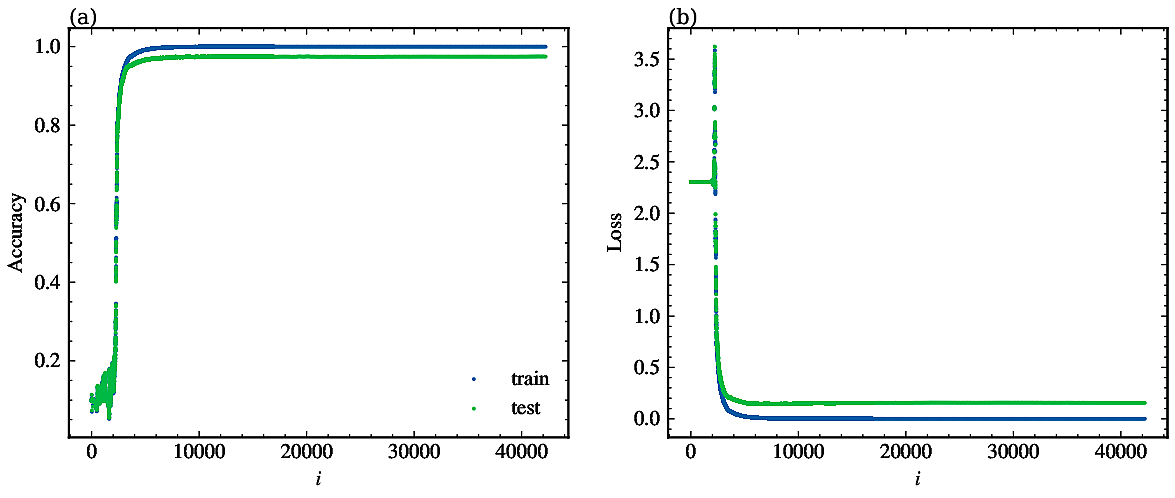}
	\caption{Accuracy of the network with weights of added Hessian eigenvectors of the entire MLP 50 network. The eigenvalues are sorted in decreasing order by magnitude. (a) Accuracy. (b) Loss.}
	\label{fig:hc_acc_abs}
\end{figure}
\begin{figure}[t]
	\centering
	\includegraphics[width=0.9\textwidth,angle=0]{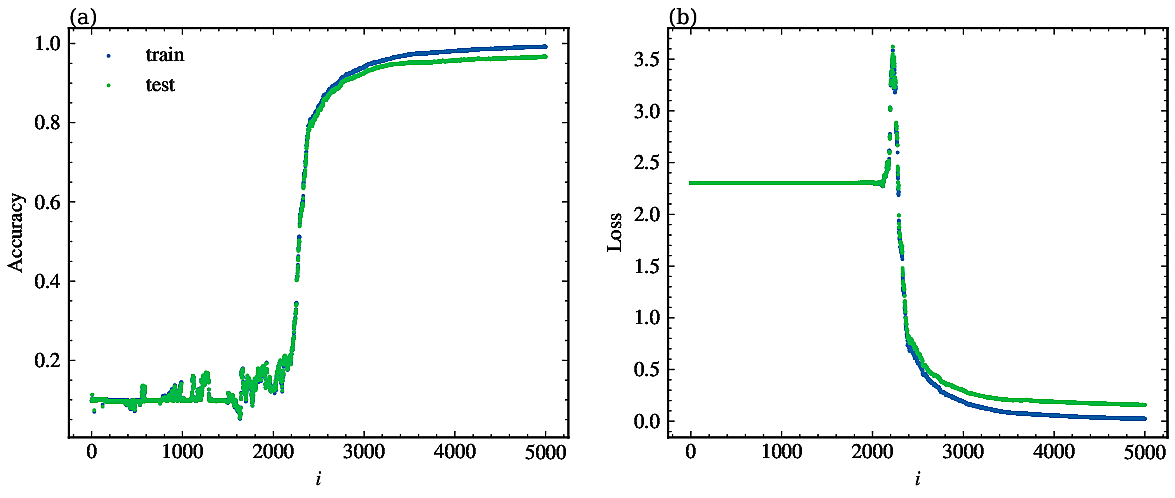}
	\caption{Accuracy of added Hessian eigenvectors of the entire MLP 50 network. The eigenvalues are sorted in decreasing order by magnitude. Zoomed in to the 5000 largest eigenvalues. (a) Accuracy. (b) Loss.}
	\label{fig:hc_acc_abs_zoom}
\end{figure}
\begin{figure}[t]
	\centering
	\includegraphics[width=0.9\textwidth,angle=0]{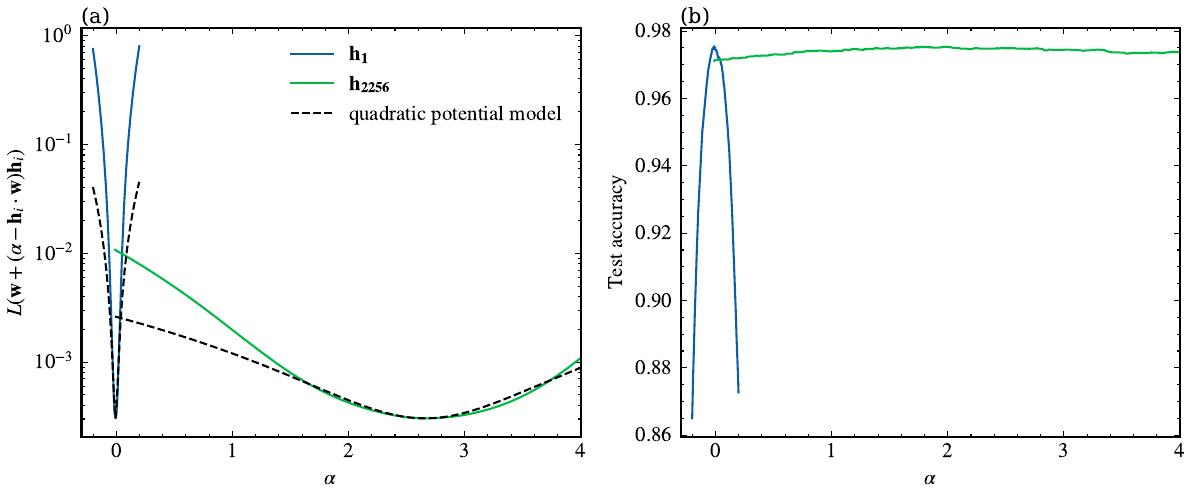}
	\caption{The loss landscape of Hessian eigenvectors of the entire MLP 50 network. The eigenvector with the largest eigenvalue and the eigenvector with the largest weight product are shown. The dashed lines represent the quadratic potential model expectation. (a) Loss. (b) Test accuracy.}
	\label{fig:potential_evh}
\end{figure}
\begin{figure}[t]
	\centering
	\includegraphics[width=0.9\textwidth,angle=0]{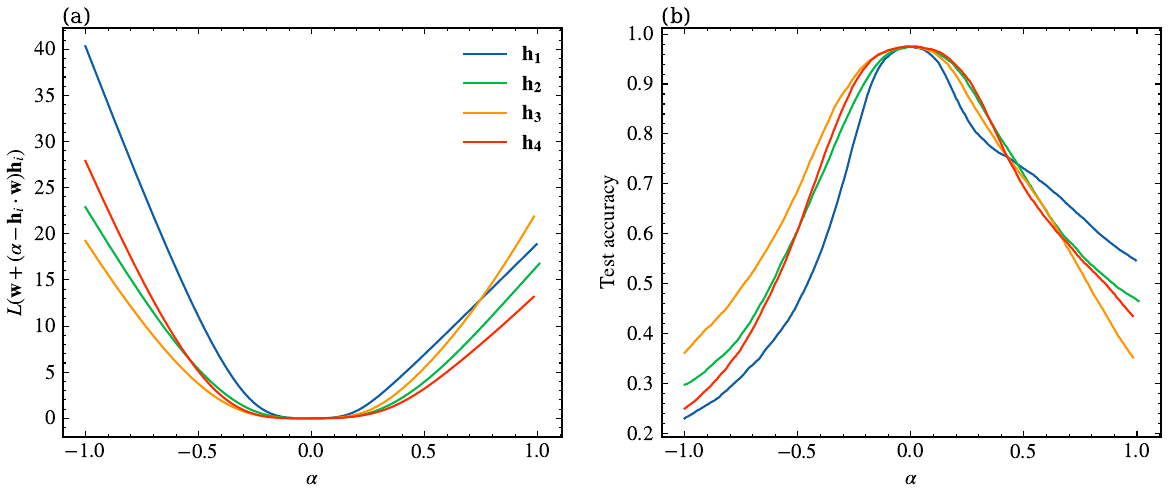}
	\caption{The loss landscape of Hessian eigenvectors of the entire MLP 50 network. The four eigenvectors with the largest eigenvalue are shown. The dashed lines represent the quadratic potential model expectation. (a) Loss. (b) Test accuracy.}
	\label{fig:potential_evh_steep}
\end{figure}
\begin{figure}[t]
	\centering
	\includegraphics[width=0.5\textwidth,angle=0]{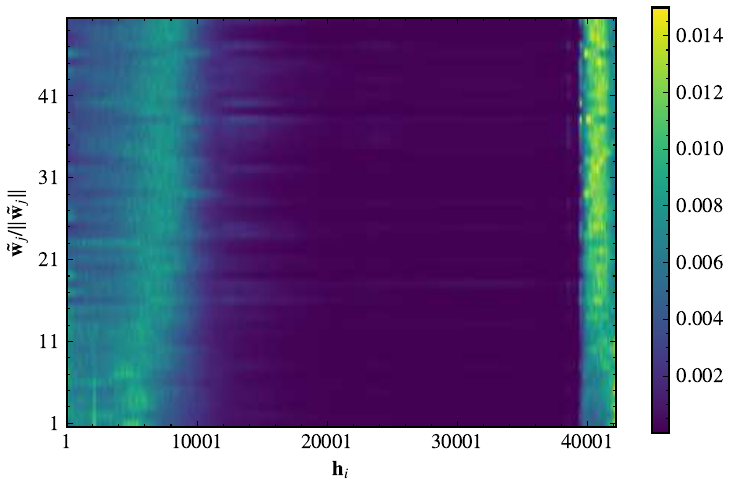}
	\caption{Singular vector product with the Hessian eigenvectors of the entire network and the singular vectors of the $(50,50)$ layer of the MLP 50 network.}
	\label{fig:sv_all}
\end{figure}

Up to this point, our focus has primarily been on analyzing the Hessian of individual layers. The underlying assumption is that each layer's Hessian must broadly exhibit the properties of the Hessian of the entire network. This assumption stems from the notion that each layer can be viewed as a network in itself, and the largest eigenvalues of each layer must align in a similar direction with the largest eigenvalues of the entire network when the eigenvectors of a layer are appended to the network's structure.\\
Let $\bm{h}_i\in \mathbb{R}^n$ represent the $i$th eigenvector of the Hessian of the whole network, and $\bm{h}^{(l)}_i\in \mathbb{R}^{n_l}$ denote the $i$th eigenvector of the Hessian of layer $l$. We can calculate:
\begin{equation}
	\label{eq:netw_layer}
	|\bm{h}_i\cdot\bm{\tilde{h}}^{(l)}_j|\, ,
\end{equation}
where $\bm{\tilde{h}}^{(l)}_j=(0,...,h^{(l)}_{j1},...,h^{(l)}_{jn_l},0,...)\in\mathbb{R}^n$.
In Fig.\ref{fig:netw_layer}, it becomes evident that, while the eigenvectors of the largest eigenvalues for both the layer and the entire network point in similar directions in the subspace, they are not identical. We can also observe a difference in all the negative eigenvalues of the total network, which fit into the picture when the absolute values are taken and the eigenvalues are resorted accordingly.\\
This observation lends credence that computing individual layer Hessian eigenvectors can be used to approximate the whole Hessian eigenvectors, given that only the largest eigenvalues need to be considered to describe the layer's behavior (as seen in Sec.\ref{sec:w_prod}). Fig.\ref{fig:network_prod_all} reveals that the weight product for the entire network exhibits behavior similar to that of individual layers for the MLP 50 network. This behavior was observed as well for the largest 12000 Hessian eigenvalues of the MLP 256 network. However, the indices with large weight products are relatively smaller in magnitude. The larger products towards the last indices correspond to negative eigenvalues that are of a magnitude similar to positive eigenvalues with the same weight product.\\
By decomposing the network into its Hessian eigenbasis and adding those together beginning with the largest eigenvalues (as in Eq.\ref{eq:adding_acc}), Fig.\ref{fig:hc_acc_nabs} shows a similar behavior as for individual layers and illustrates that the eigenvectors of negative eigenvalues are necessary to achieve the network's full performance. Consequently, an alternative approach could involve sorting eigenvalues in decreasing order by absolute size. The accuracy for this sorting scheme is depicted in Fig.\ref{fig:hc_acc_abs}, where the network achieves full performance qualitatively similar to the behavior of individual layers, with approximately $\approx 2500$ added eigenvectors. In Fig.\ref{fig:hc_acc_abs_zoom} we can see that the accuracy start to improve not earlier than for about the $2000$th eigenvalue and that during this transition the loss increases instead.\\
To further comprehend the significance of directions with large weight products for accuracy, we can inspect the loss landscape with:
\begin{equation}
	L\left(\bm{w}+(\alpha-\bm{w}\cdot\bm{h}_i)\bm{h}_i\right)\, .
\end{equation}
For $\alpha=0$ we remove the $i$th eigenvector completely from the network and for $\alpha=\bm{w}\cdot\bm{h}_i$ it is fully included. Fig.\ref{fig:potential_evh} demonstrates that the potential well corresponding to the larger eigenvalue is indeed steeper, yet so close to zero that its importance cannot be shown by simply setting its projection to zero. For both potentials, it is observable that the quadratic approximation holds only for $\alpha-\bm{w}\cdot\bm{h}_i\lesssim 0.1$. Setting the projection of the eigenvector with the largest weight product to zero takes us far from the quadratic approximation. While the loss prediction becomes less accurate, the qualitative description of a significantly larger loss remains valid.
We observe from Fig.\ref{fig:potential_evh_steep} that the hypothesis stating that the minimum of the largest eigenvalues are in proximity to zero holds.\\
Using the flattened singular matrices, we can once again compare them to the Hessian eigenvectors, similar to the layer and network eigenvector comparison. Fig.\ref{fig:sv_all} reveals that only the $12000$ largest eigenvalues in magnitude have a relatively large product to the flattened singular matrices that make them distinct from smaller eigenvalues. Outlying singular values exhibit significant scalar products with eigenvectors of smaller indices, while singular values within the bulk display larger scalar products with eigenvectors of larger indices.

\section{Distribution of the Hessian Eigendecomposition}
\label{sec:dist}
\subsection*{Distribution of Eigenvalues}
\begin{figure}[t]
	\centering
	\includegraphics[width=0.5\textwidth,angle=0]{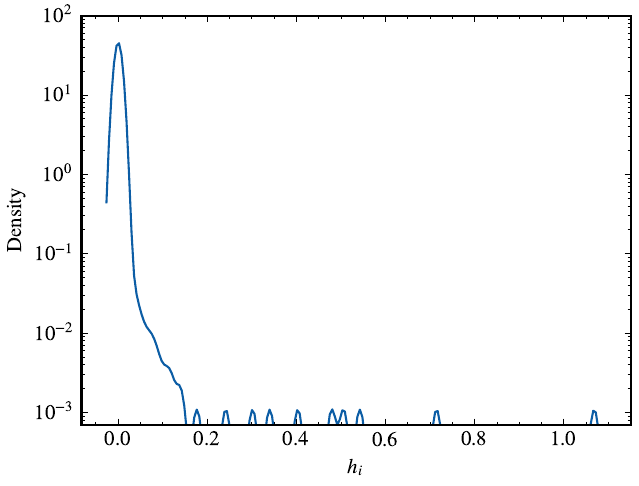}
	\caption{Spectral density of the Hessian eigenvalues of the entire MLP 50 network. The standard deviation of the smoothing kernel is set to one.}
	\label{fig:kde_hess}
\end{figure}
\begin{figure}[t]
	\centering
	\includegraphics[width=0.5\textwidth,angle=0]{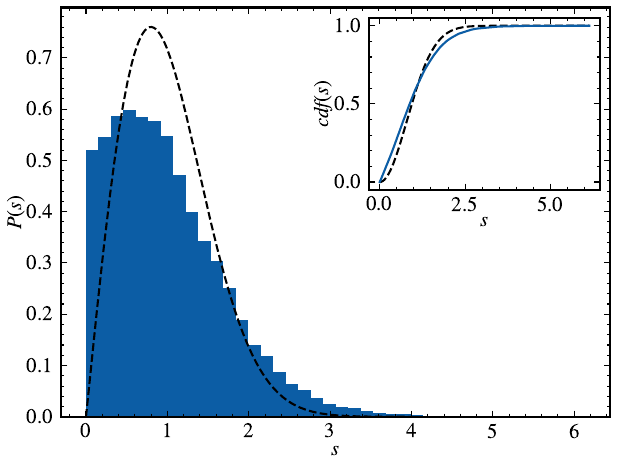}
	\caption{Level spacings of the unfolded spectrum of Hessian eigenvalues of the entire MLP 50 network. The dashed line represents the expectation for random matrices of the same size. Averaged over the $10$ nearest neighbors in both directions. The inset plot shows the cumulative distribution function. Adapted from \cite{Thamm_2022}.}
	\label{fig:wig_sum}
\end{figure}

Examining the spectral distribution of the Hessian eigenvalues in Fig.\ref{fig:kde_hess}, it is apparent that the largest eigenvalues greatly surpass the bulk of eigenvalues in magnitude, and a small fraction of eigenvalues are negative. The introduction of weight decay alters the Hessian matrix to be summed with $2\lambda\mathbb{1}_{n\times n}$, leading to a shift of all eigenvalues by $2\lambda$. Consequently, the primary concentration of eigenvalues shifts from zero to $2\lambda$. Weight decay change the location of the minimum and the minimum found by the network during training.
The spacing of eigenvalues does not conform to the Wigner surmise, as evidenced by Fig.\ref{fig:wig_sum}. Suggesting that the Hessian eigenvalues are not randomly distributed.

\subsection*{Properties of Eigenvectors}
\begin{figure}[t]
	\centering
	\includegraphics[width=0.5\textwidth,angle=0]{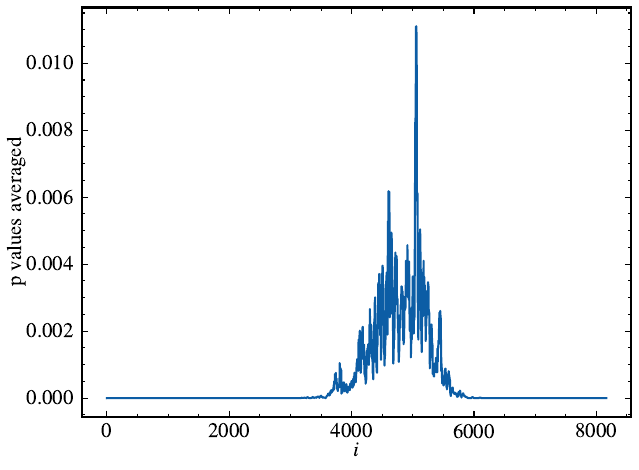}
	\caption{Averaged p-values of the $(128,64)$ layer of the unregularized MLP 256 network. Averaged over the $15$ closest eigenvalues in both directions.}
	\label{fig:ptd}
\end{figure}

The Porter-Thomas distribution \cite{porter1956fluctuations} predicts that a vector $\bm{v}$ with random entries, having its entries sorted in increasing order $\tilde{\bm{v}}$, we have that 
\begin{equation}
	1+\frac{\text{erf}(\tilde{v}_i\sqrt{\frac{N}{2}})}{2}\approx i/N\, ,
\end{equation}
where $\text{erf}$ is the error function and $N$ the length of the vector.
Assessment against the Porter-Thomas distribution, reveals that none of the eigenvectors conform to this random distribution. A p-value near $0.5$ would indicate that the eigenvectors follow the distribution, but numerically, all p-values are effectively zero. This indicates that the eigenvectors deviate substantially from the distribution.\\
This conclusion holds for the Hessian of the entire MLP 50 network, and for the eigenvalues and eigenvectors of the Hessian of individual layers with $4096 - 42310$ parameters of all other networks as discussed in Sec.\ref{sec:Networks}. In instances where layers have few parameters, the p-values may reach approximately $\approx 0.02$ at most when averaged over the p-values of eigenvectors from local groups, as shown in Fig.\ref{fig:ptd}. Larger p-values might be attributed to eigenvectors having fewer parameters or being closer to random than the eigenvectors of the whole network.

\subsection*{Layerwise Concentration of Eigenvectors}
\begin{figure}[t]
	\centering
	\includegraphics[width=0.5\textwidth,angle=0]{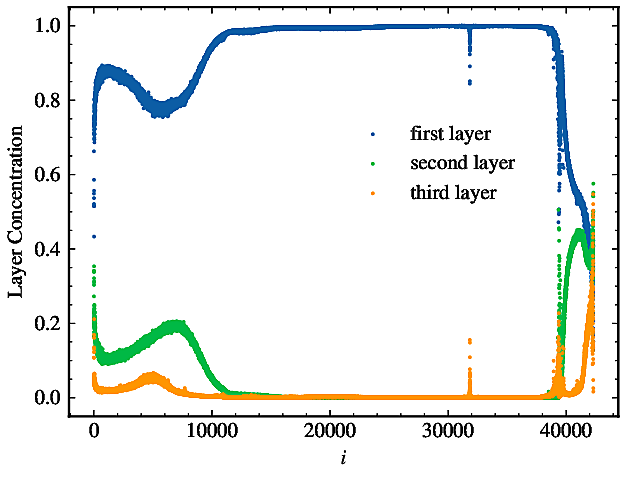}
	\caption{Concentration of Hessian eigenvectors in specific layers by the square norm of each layer's components. The MLP 50 network is analyzed for this illustration.}
	\label{fig:localization}
\end{figure}

Considering the square of the norm for sections of total eigenvectors:
\begin{equation}
	\label{eq:square_norm}
	\|\bm{h}_i\|^2_l=\sum_{j\in I_l}h^2_{ij}\, ,
\end{equation}
where $I_l$ represents the interval of indices within layer $l$, Fig.\ref{fig:localization} demonstrates that for most small eigenvalues, localization predominantly resides in the first layer. Assuming vector entries are completely random, the expected concentration fraction for each layer would correspond to the ratio of its number of parameters to the total number of parameters. In the provided figure, the first layer is anticipated to have a fraction of $\approx 0.928$, the second $\approx 0.06$, and the third $\approx 0.012$. This comparison highlights that eigenvectors of the largest eigenvalues are more concentrated in later layers. One plausible explanation for this concentration lies in the vanishing gradient problem, layers farther from the output layer tend to possess smaller gradients due to the multiplication of gradients from subsequent layers. This training characteristic results in more attention directed toward later layers, which can be expected to possess steeper minima, leading to larger Hessian eigenvalues.

\section{Development of the Network in the Direction of Hessian Eigenvectors}
\label{sec:dev}
\begin{figure}[t]
	\centering
	\includegraphics[width=0.9\textwidth,angle=0]{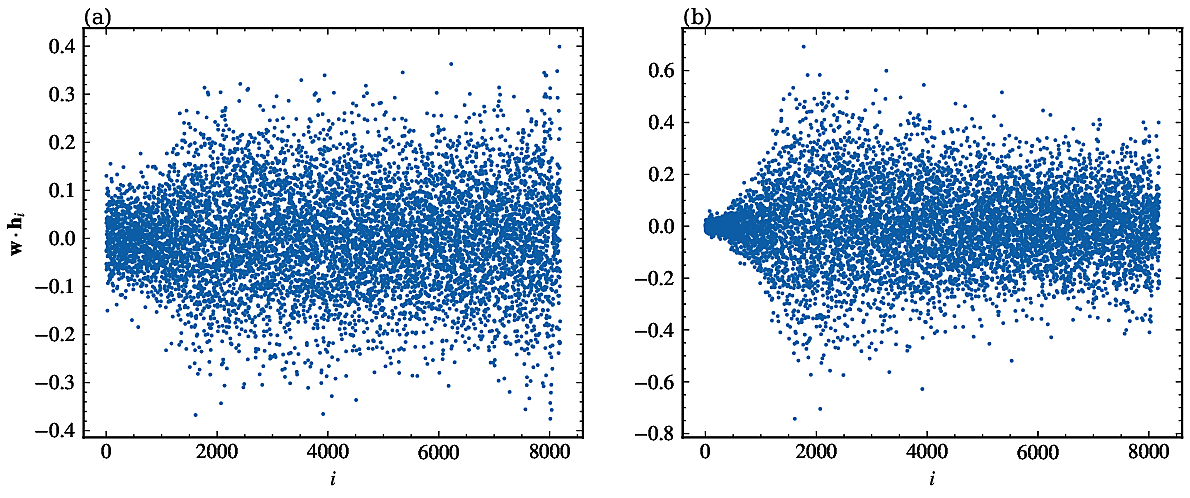}
	\caption{Weight product of the $(128,64)$ layer of the unregularized MLP 256 network. (a) At network initialization. (b) After 100 epochs of training. Eigenvectors are computed after 100 epochs of training for both figures.}
	\label{fig:weights_prod_init}
\end{figure}

\begin{figure}[t]
	\centering
	\includegraphics[width=0.9\textwidth,angle=0]{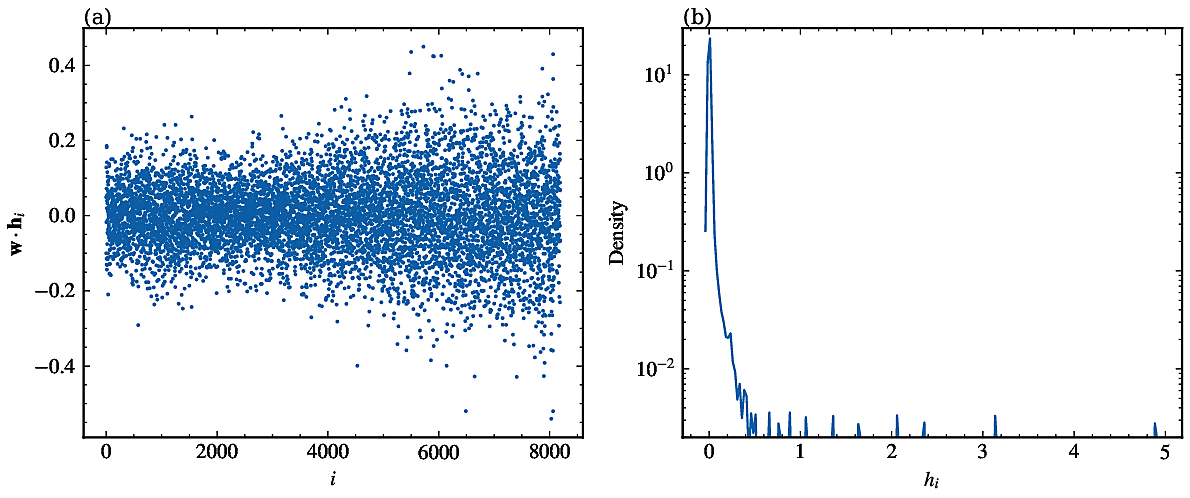}
	\caption{Hessian eigenvalues and eigenvectors of the $(128,64)$ layer of the unregularized MLP 256 network at initialization. (a) Weight product. (b) Spectral density. The standard deviation of the smoothing kernel is set to one.}
	\label{fig:weights_prod_kde}
\end{figure}

\begin{figure}[t]
	\centering
	\includegraphics[width=0.9\textwidth,angle=0]{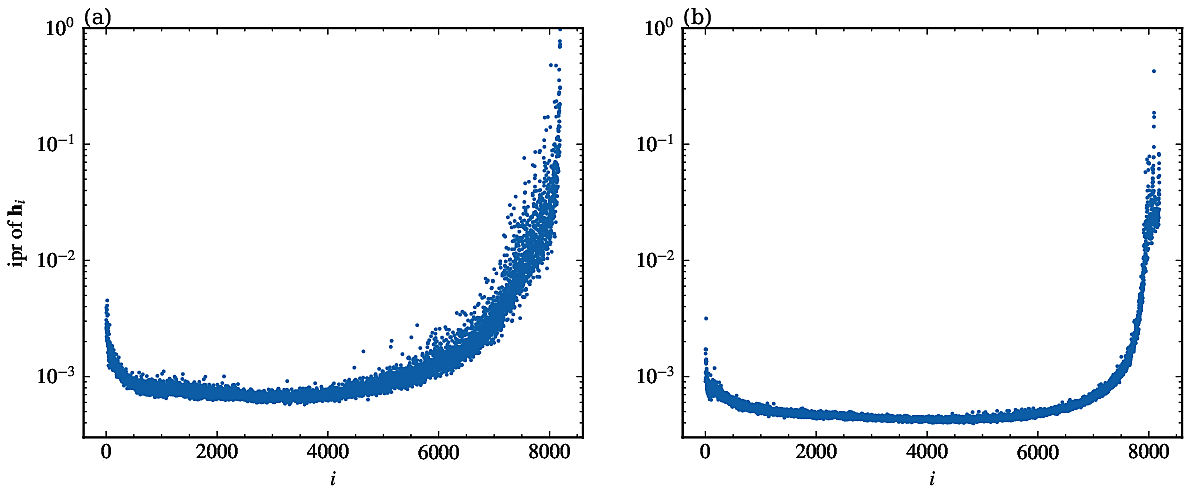}
	\caption{Inverse participation ratio of the $(128,64)$ layer of the unregularized MLP 256 network. (a) At initialization.  (b) After 100 epochs.}
	\label{fig:ipr}
\end{figure}

As observed in Sec.\ref{sec:w_prod}, the scalar product of Hessian eigenvectors and network weights demonstrates structured behavior. To gain insight into this behavior, we examine how eigenvectors and weights evolve during training, not only during the exploration phase but also during initialization.\\
Fig.\ref{fig:weights_prod_init} indicates that even at initialization, the weight product for larger Hessian eigenvalues is smaller than for smaller Hessian eigenvalues, that were computed after training. This suggests that network initialization contributes to determining the final network shape, despite employing random initialization.\\
In Fig.\ref{fig:weights_prod_kde} the eigenvalues are already shared a similar spectrum as to Fig.\ref{fig:kde_hess} before training, with a few large eigenvalues and numerous others clustered around zero. This suggests that the dataset and the structure of the network significantly influence the distribution of Hessian eigenvalues, as they are the only contributions of the network not initialized randomly. Although the weight product for these eigenvalues before training seems more randomly distributed, it yields smaller products for larger eigenvalues than for smaller ones.\\
Fig.\ref{fig:ipr} illustrates that the inverse participation ratio $\text{ipr}(\bm{h}_i)=\sum_jh^4_{ij}$ is large for eigenvectors with the largest and smallest eigenvalues, while it is notably smaller for eigenvectors in between. However, it remains localized, in comparison to the full delocalization of $1/n=1/8192\approx0.0001$ for the considered layer, where each component has a value of $1/\sqrt{n}$.\\
Let us assume that weights at initialization follow a random distribution with zero mean and variance $\sigma^2$, while eigenvector components possess zero mean and variance $\sigma^2_i$, with both variables assumed independent. Accordingly:
\begin{equation}
	\langle\left(\bm{w}\cdot\bm{h}_i\right)^2\rangle=\sum_j\langle w^2_j\rangle\langle h^2_{ij}\rangle=\sigma^2\sigma^2_{i}=\sigma^2\, ,
\end{equation}
where we leverage eigenvector normalization, i.e., $1=\langle\|\bm{h}_i\|^2\rangle=\sum_j\langle h^2_{ij}\rangle=\sigma^2_i$.
Hence, the eigenvectors and weights should not be assumed to be independent, as the distribution of the weight product would then not depend on the index.

\section{Catastrophic Forgetting}
\label{sec:eva_cata_forget}
\begin{figure}[t]
	\centering
	\includegraphics[width=0.5\textwidth,angle=0]{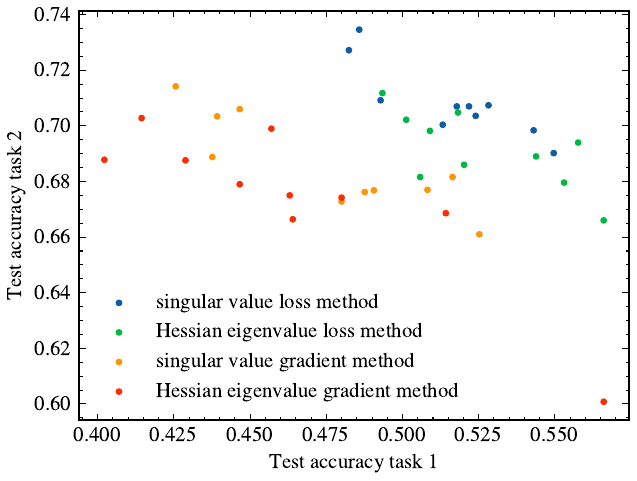}
	\caption{Catastrophic forgetting accuracies on the LeNet, where only the last layer is updated. Each dot of the same color represents the result of another seed. The maximum sum of both accuracies is taken for each seed.}
	\label{fig:cf_last}
\end{figure}

\begin{figure}[t]
	\centering
	\includegraphics[width=0.9\textwidth,angle=0]{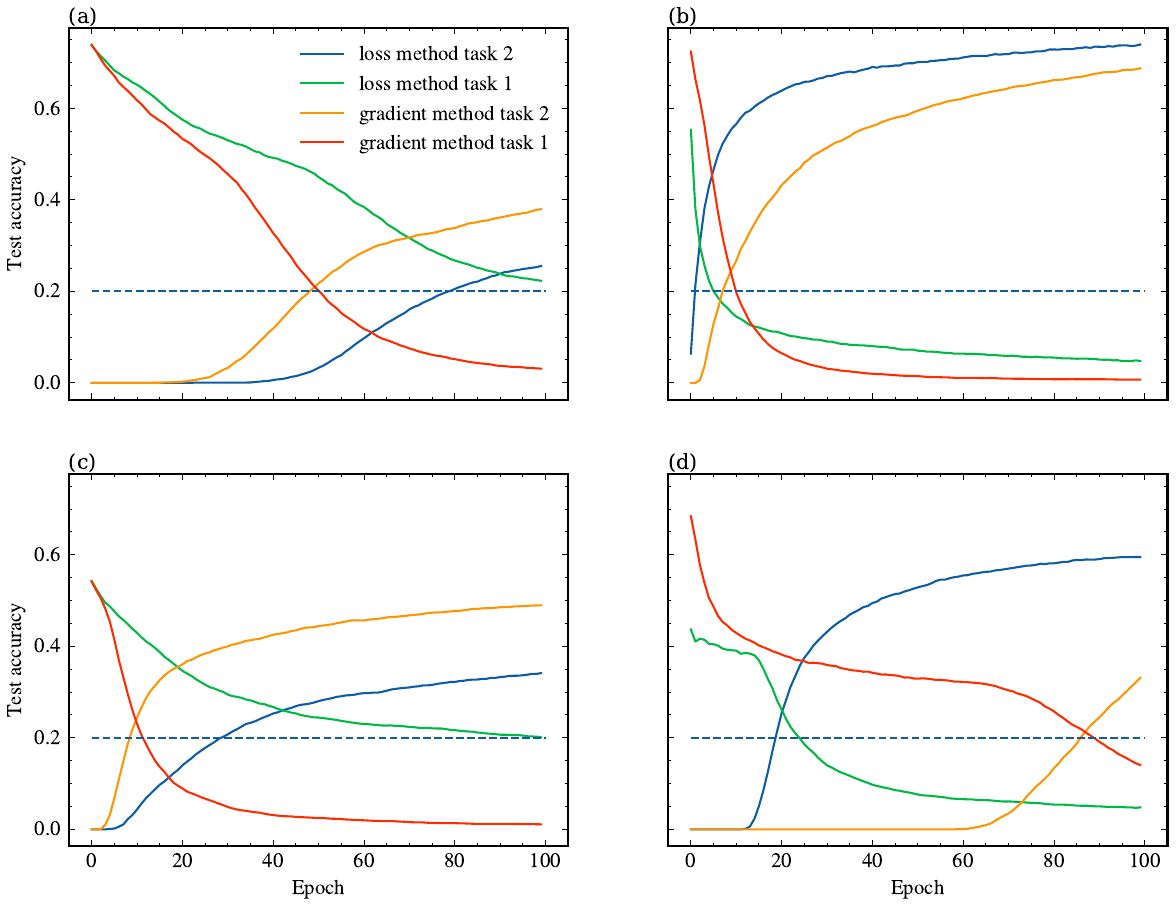}
	\caption{Catastrophic forgetting accuracies for updating all layers for the second task during training. (a) LeNet. (b) Mini AlexNet. (c) MLP 256 without weight decay. (d) ResNet 20. The dashed line refers to random guessing of both individual tasks.}
	\label{fig:cf_all}
\end{figure}

In Sec.\ref{sec:cata_forget}, we proposed, following the idea of \cite{doi:10.1073/pnas.2015617118}, that preserving the larger Hessian eigenvalues can potentially lead to maintaining test accuracies for other tasks in terms of multiple task training. This was accomplished by incorporating the regularization term:
\begin{equation}
	L_\text{cf}(\bm{w}) = \lambda_{cf}\sum^{N_\text{lim}}_{i=1}h_i\left((\bm{w}-\tilde{\bm{\mu}}_1)\cdot\bm{h}_i\right)^2
\end{equation}
into the loss function. Here, $\lambda_{cf}$ represents a positive real regularization constant, $\tilde{\bm{\mu}}_1$ denotes the weights, $h_i$ signifies Hessian eigenvalues in descending order, and $\bm{h}_i$ corresponds to their associated eigenvector, all at the end of training for the previous task. Building upon the insights from Sec.\ref{sec:w_prod} and Sec.\ref{sec:svhess}, we propose a novel regularization term:
\begin{equation}
	\label{eq:cf_sv}
	L_\text{sv}(\bm{w}) = \lambda_{\text{sv}}\sum^{N_\text{out}}_{i=1}\nu_i\left((\bm{w}-\tilde{\bm{\mu}}_1)\cdot\frac{\tilde{\bm{w}}_i}{\|\tilde{\bm{w}}_i\|}\right)^2\, ,
\end{equation}
where $\lambda_\text{sv}$ is a regularization constant, $N_\text{out}$ denotes the number of singular values outside the RMT bulk, $\nu_i$ are the singular values, $\tilde{\bm{w}}_i$ are the corresponding flattened singular matrices, where the diagonal matrix had all entries set to zero except the $i$th singular value, and $\tilde{\bm{\mu}}_1$ is the weight vector after training for the first task. This regularization strategy brings two computational advantages. Firstly, computing the singular values of a weight matrix is significantly less demanding in terms of memory and computation compared to the computation of Hessian eigenvalues. Secondly, during each training step where $\nabla_{\bm{w}}L_\text{sv}$ needs to be computed, the sum is performed over far fewer indices compared to $N_\text{lim}$, as we can expect $N_\text{out} \ll N_\text{lim}$.\\
For biases, this regularization procedure is not directly applicable, but due to their relatively small vectors in comparison to the weight matrices of layers, their Hessian matrices can be computed efficiently and thus regularized similarly to the approach of the Hessian eigenvalues.\\
To empirically compare the effectiveness of our proposed singular value approach against the Hessian eigenvalue method, we adopt a strategy similar to that detailed in \cite{doi:10.1073/pnas.2015617118}. The experimental setup involves training a network to attain full training accuracy across all tasks at once. This entails employing pre-trained networks and reinitializing the layers under investigation for our measurement. Specifically, we focus on the layers relevant to our analysis and train it exclusively to achieve full training accuracy for the first task. Subsequently, we extract singular values and Hessian eigenvalues from this process and introduce the corresponding regularization component to the layers.\\
By then training the modified layers on the second task, we can assess the impact on the first task's performance. Without regularization, this can lead to a significant decrease in performance, potentially plummeting to levels comparable to random guessing and lower. This is due to the network's inclination to predict labels for the second task, thereby undermining its competence in the first task. Notably, the first task involves classifying the first five label categories, while the second task involves classifying the remaining five label categories from the CIFAR-10 dataset. We further train the network for 50 epochs using each method, selecting the results of the epoch with the highest sum of accuracies for both tasks. Both methods employ a substantial regularization constant of $\lambda_\text{cf} = 1000$. However, higher constants could potentially lead to numerical instability after a single training step, depending on the learning rate. The learning rate here is set to $1\%$ of the final learning rate from the preceding training phase. Lower learning rates favor the preservation of the initial task's accuracy, though at the cost of requiring more epochs to achieve acceptable accuracy for the second task. For the singular value method, we conserve either $20\%$ of the Hessian eigenvalues of the biases or the five largest singular values of matrices and higher order tensors. While preserving weights based on previous tasks might seem unconventional, since the solution we fix the regularization on might not be the optimal solution for both tasks, an alternative approach is to directly impose the constraint on gradients rather than loss. We here propose the equation:
\begin{equation}
	\bm{\tilde{v}}(t):=\bm{v}(t)-\gamma\sum^{N_{out}}_{i=1}\frac{\nu_i}{\nu_1}(\bm{v}\cdot\tilde{\bm{w}}_i)\tilde{\bm{w}}_i\, ,
\end{equation}
where $\bm{v}(t)$ is the velocity of the unregularized network, and $\gamma \in [0,1]$ represents the constraint strength. When $\gamma=1$, the equation effectively prevents weight updates from occurring in conserved directions. The fraction of singular values in the equation ensures that the largest singular value and singular matrix is fully conserved, while smaller ones are conserved partially. A similar fraction is employed for Hessian eigenvalues. Importantly, this gradient-constraint approach circumvents numerical instability, rendering it a more stable option.\\
Fig.\ref{fig:cf_last} presents the results for both gradient and loss-based methods. When comparing the sum of test accuracies, the former regularization method directly conserving weights yields superior overall performance. Intriguingly, both the singular value and Hessian methods yield comparable regularization performance, regardless of the fact that the singular value method conserves fewer directions. Consistent results as in the figure are achieved when performing separate computations for each layer individually, including convolutional layers.\\
However, the desire to preserve all layers simultaneously leads to a significant decrease in overall performance, as seen in Fig.\ref{fig:cf_all}. The maximal overall test accuracy achieved using the regularization singular value loss method on the whole LeNet is $35\%$, while the gradient singular value approach results in a maximum test accuracy of $25\%$, still $5\%$ higher than random guessing for the five-class tasks. It is important to note that due to training on two tasks, the limit for random guessing is no longer $20\%$ for each task, but $0\%$, as the network could classify all images into labels from the other task. The minimum of both tasks averaged is $10\%$, similar to the case of training for all labels simultaneously. This interpretation allows us to see these accuracies as not entirely incorrect but slightly lower compared to training without tasks.\\
An explanation for the better performance when training only a few layers could be that when the network is initially trained on all tasks, it might seek a solution that approximates solving all tasks simultaneously. However, preserving a layer could guide the network towards a solution more similar to the pre-reinitialization solution. By training the network exclusively on one task, the need for compatibility with the other task's solution becomes unnecessary. This was further investigated by training multiple layers simultaneously to assess the accuracy of this assumption. The obtained accuracies remain comparable to those in Fig.\ref{fig:cf_last} when training layers with fewer parameters. This suggests that the improvement in accuracy in some layers could be attributed to the fixation of other layers to stable solutions. When training multiple layers, the Hessian eigenvector regularization method seems to outperform the singular value method by up to $2\%$. This result might be due to the singular value method not fully covering the weight space. Alternatively, there could be room for refining the optimization of regularization constraints, an aspect that warrants further investigation.

\chapter{Discussion}
\label{Discussion}

\section{Generalizability of our Results}
In this section, we discuss into the generality of our results concerning network structures that were not explicitly included in our analysis. It is worth noting that the influence of the update rule and batch size primarily manifests itself in the context of principal component analysis, since it is a dynamical property. Conversely, the Hessian matrix and singular value decomposition remain relatively invariant with respect to these properties. Their influence, if any, operates indirectly through the minima that they help identify.
\subsection*{Momentum}
The impact of momentum in early training is notable, as it can influence the minima discovered by the network. When momentum $\beta$ is applied during the exploration phase, it alters the effective learning rate, assuming a near-constant gradient. Utilizing the SGD equation $\bm{v}(t+1)=-\eta\nabla_{\bm{w}}l^{(k(t))}(\bm{w},\bm{x},\bm{z})+\beta\bm{v}(t)$, where $\eta$ is the learning rate, $l^{(k(t))}$ the loss of batch $\mathcal{B}_k$ and $\bm{w}(t+1)=\bm{w}(t)+\bm{v}(t+1)$, we can deduce:
\begin{equation}
	\begin{split}
		\langle \bm{v}(t)\rangle_\mathcal{B}
		&=-\eta\sum^t_{t'=0}\beta^{t-t'}\langle\nabla_{\bm{w}}l^{(k(t'))}(\bm{w},\bm{x},\bm{z})\rangle_\mathcal{B}
		\\&=-\frac{\eta}{S} \nabla_{\bm{w}} L \sum^t_{t'=0}\beta^{t-t'}
		=-\frac{\eta}{S} \nabla_{\bm{w}} L \sum^t_{t'=0}\beta^{t'}
		\\&=-\frac{\eta}{S} \frac{1-\beta^{t+1}}{1-\beta} \nabla_{\bm{w}} L
		\\&\approx -\frac{\eta}{S(1-\beta)} \nabla_{\bm{w}} L\, ,
	\end{split}
\end{equation}
where we used the geometric series formula and that $\langle\nabla_{\bm{w}}l^{(k)}(\bm{w},\bm{x},\bm{z})\rangle_\mathcal{B}=\frac{1}{S}\nabla_{\bm{w}} L$ is the gradient of the mean loss of all batches. Momentum renormalizes the bare learning rate to become effectively larger. Excluding this, momentum's presence does not alter the observed qualitative behavior for the PCA, the findings indicate the generalizability of these observations to various minima that can be approximated by the network utilizing momentum.

\subsection*{Variants of SGD}
While momentum's effect on the minimum in our thesis is understood, the effect under other SGD variants like Adam \cite{kingma2017adam} and Lion \cite{chen2023symbolic} remains unexplored in our thesis. The dynamics of update steps for these methods are more complex. However, it is anticipated that network parameters may tend to converge to flat minima, resembling those of SGD. This commonality could lead to similar qualitative behaviors observed with SGD.

\subsection*{Batch Size}
This work has not explicitly addressed the impact of batch size. Qualitative results have been validated for typical batch sizes within the range of $S \in [32, 512]$ for SGD. Altering the batch size influences the minima found by the network \cite{jastrzębski2018factors}. Larger batch sizes have been found to yield steeper loss landscapes, corresponding to larger eigenvalues in the Hessian matrix \cite{jastrzębski2018factors}.

\subsection*{Layer and Network Analysis}
Analyzing network weights can be simplified by considering subsets of layers, treating their weights of separate layers as multiple independent parameters. Hessian matrix analysis is more intricate, as the loss gradient of different layers is interconnected. Despite this complexity, results in Sec.\ref{sec:full_hess} show that the Hessian eigenvectors for individual layers closely resemble the corresponding eigenvectors of the entire network of similar indices. This implies that for large eigenvalues, computing Hessian eigenvectors for individual layers suffices, making such computations feasible even for large networks with layers having no more than $\lesssim 50000$ parameters, requiring around $\approx 128 \,\text{GB}$ of RAM.

\section{Further Questions}

\subsection*{Data Augmentation}
The importance of data augmentation in modern networks is well-recognized \cite{he2016deep}. However, incorporating data augmentation into Hessian matrix evaluation poses challenges due to the vast number of possible realizations of training samples. The more training examples a network uses during training, the more costly the computation of the total loss is, since the Hessian must be computed for each sample separately. To approximate the total loss, a sufficiently large number of augmented samples must be used, mimicking the strategy of approximating generalization performance with a smaller test dataset compared to the training dataset.

\subsection*{Batch Normalization}
Batch normalization's influence on training dynamics is substantial. This creates a correlation between the samples in a batch \cite{ioffe2015batch}. Without batch normalization, samples could be assumed to be independent with respect to their gradient. For our ResNet 20, the influence of batch normalization did not change the qualitative behavior in the weight product, but quantitatively.

\subsection*{Noise Filtering}
Recent work has suggested noise filtering in the context of label noise \cite{Staats.2022}. Removing the small singular values from the network and reshaping the distribution to align with a noise-free scenario has shown to improve test accuracy. One explanation is that these singular values of the RMT bulk are not changed by the network during training, or are changed so little that they do not contain true information, but only noise, such as label noise caused by misclassification. This can result in a shift of the RMT bulk towards larger singular values, which is small enough for a realistic amount of label noise $\leq 40\%$, such that outlying singular values can still be separated from the ones of the bulk \cite{Staats.2022}. Using our observations and relationships to the Hessian eigenvectors, we can further extend our understanding. Recall that the eigenvectors of the largest Hessian eigenvalues have a large product with the singular matrices of the largest singular values as shown in Sec.\ref{sec:svhess}. Thus, if we set small singular values to zero, they will mostly affect only the Hessian eigenvectors of the small eigenvalues, which we know by setting them to zero have such flat potentials that shifting them cannot measurably increase the loss seen in Fig.\ref{fig:potential_evh}. In this work it was not covered how label noise affects the Hessian eigenvalues and its properties to the weights. This can be done in future work to the extent of our understanding of handling noise practically.

\subsection*{Information Transition of the Dataset}
Sec.\ref{sec:dev} shed light on the remarkable alignment between the pre-training and post-training eigenvalue orders, highlighting the influence of both the dataset's characteristics and the network's architecture on the observed patterns. This insight offers a complement to dataset analyses, as explored in \cite{levi2023underlying}.

\subsection*{Catastrophic Forgetting}
The proposed catastrophic forgetting regularization  in Sec.\ref{sec:eva_cata_forget} performs well on individual layers but less effectively on entire networks. While the method applied to the gradient seems to fail completely in terms of test accuracy, the loss method performs reasonably well and may be applicable. Our results may be even more useful for fine-tuning a network, i.e., training on the latest layers to further specialize the network, e.g., training the network on a task it has never seen before \cite{sanh2021multitask}. This can be few-shot learning \cite{liu2022few}, where the network has to recognize label classes it has never seen before that are very similar to those it has already learned. The CIFAR-10 dataset we used to test catastrophic forgetting does not have enough label classes to efficiently predict how our method will perform on numerous label classes. Considering that as problems get harder, the number of label classes becomes so large that many classes are only seen by the network after many batches, preventing catastrophic forgetting may become more important for larger classification problems.

\subsection*{Explainability of Deep Neural Networks}
This work represents a stride towards unraveling the enigmatic nature of Deep Neural Networks (DNNs), contributing to their explainability. Our understanding provides a lens through which we can scrutinize and interpret the behavior and properties of DNNs. Furthermore, this pursuit holds the potential to assuage growing societal concerns about the evolving capabilities of these networks, as emphasized in recent discussions \cite{HEUILLET2021106685,metz2023}.

\chapter{Conclusion}
\label{conclusion}
In this study, we delved into the intricate dynamics of trained neural networks, unraveling key insights that shed light on their behavior. Trained networks predominantly continue training in a single direction, known as the drift mode. This intriguing drift mode can be elegantly explained by the quadratic potential model of the loss function, suggesting a slow exponential decay towards the potential minima.\\
Through our analysis, we unveiled a strong correlation between the Hessian eigenvectors and the network weights. This relationship, hinged on the magnitude of eigenvalues, allowed us to discern the important parameter directions within the network. Notably, the significance of these directions rests on two defining attributes: the curvature of their potential wells (indicated by the magnitude of Hessian eigenvalues) and their alignment with the weight vectors.\\
Our exploration extended to the decomposition of weight matrices through singular value decomposition. Quite surprisingly the overlap of Hessian eigenvectors of large eigenvalues is larger for singular matrices of larger singular values than for the weight vector.\\ Furthermore, our examination showcased the applicability of principal component analysis in approximating the Hessian, with update parameters emerging as a superior choice over weights for this purpose.\\
Remarkably, our findings unveiled a similarity between the largest Hessian eigenvalues of individual layers and the entire network. Notably, higher eigenvalues were concentrated more in deeper layers. Strikingly, the Hessian eigenspectrum, far from being randomly distributed, exhibited a pronounced pattern predetermined by the dataset and the network structure, even before the onset of training.\\
Leveraging these insights, we ventured into the realm of addressing catastrophic forgetting a challenge that plagues neural networks when learning new tasks while retaining knowledge from previous ones. By applying our discoveries of the overlaps of the Hessian eigenvectors and singular matrices, we formulated an effective strategy to mitigate catastrophic forgetting, offering a pragmatic solution that can be applied to networks of varying scales, including larger architectures.\\
In conclusion, our journey through the intricate landscapes of trained neural networks has revelations that deepen our understanding of their behavior. These insights hold the promise of not only refining network training but also influencing the broader discourse on explainability and reliability in the realm of deep learning.

\singlespacing

\newpage
\phantomsection
\addcontentsline{toc}{chapter}{Bibliography}

\printbibliography

\onehalfspacing

\newpage
\phantomsection

\end{document}